\documentclass[review]{elsarticle}

\usepackage{graphicx}
\usepackage{amssymb}
\usepackage{amsthm}
\usepackage{amsmath}
\usepackage{bm}
\usepackage{color}
\usepackage{tabularx} % for tables
\usepackage{booktabs} % for rulers
\usepackage{siunitx} % for si units in tables
\usepackage[caption=false]{subfig} % for subfloats
\usepackage{capt-of}
\usepackage{natbib}
\usepackage{float}

\usepackage{pgfplots} 
\pgfplotsset{compat=1.16}

\newcommand{\condbar}{\,|\,}  % The vertical bar in f(y | x)
\newcommand{\reals}{\mathbb{R}}

\newcommand{\expectationp}[2][]{\mathbf{E} \ifx #1 \undefined \else _{#1} \fi \left[#2\right]}
\usepackage[linktocpage]{hyperref} 			% Load hyperref as last to avoid compatibility issues. This option works with latex.

\biboptions{authoryear}

\usepackage{fancyhdr}

\begin{document}

\begin{frontmatter}
\title{On gray-box modeling for virtual flow metering}

\address[label1]{Department of Engineering Cybernetics, NTNU, O. S. Bragstads plass 2D, 7034 Trondheim, Norway}
\address[label2]{Solution Seeker AS, Rådhusgata 24, Oslo, Norway}
\address[label3]{TechnipFMC, Philip Pedersens vei 7, Lysaker, Norway}
\cortext[cor1]{I am corresponding author}

\author[label1,label2]{Mathilde Hotvedt\corref{cor1}}
\ead{mathilde.hotvedt@ntnu.no}

\author[label1,label2]{Bjarne Grimstad}
\ead{bjarne.grimstad@solutionseeker.no}

\author[label3]{Dag Ljungquist}
\ead{Dag.Ljungquist@technipfmc.com}

\author[label1]{Lars Imsland}
\ead{lars.imsland@ntnu.no}

\begin{abstract}
A virtual flow meter (VFM) enables continuous prediction of flow rates in petroleum production systems. The predicted flow rates may aid the daily control and optimization of a petroleum asset. Gray-box modeling is an approach that combines mechanistic and data-driven modeling. The objective is to create a computationally feasible VFM for use in real-time applications, with high prediction accuracy and scientifically consistent behavior. This article investigates five different gray-box model types in an industrial case study using real, historical production data from 10 petroleum wells, spanning at most four years of production. The results are diverse with an oil flow rate prediction error in the range of 1.8\%-40.6\%. Further, the study casts light upon the nontrivial task of balancing learning from both physics and data. Therefore, providing general recommendations towards the suitability of different hybrid models is challenging. Nevertheless, the results are promising and indicate that gray-box VFMs can reduce the prediction error of a mechanistic VFM while remaining scientifically consistent. The findings motivate further experimentation with gray-box VFM models and suggest several future research directions to improve upon the performance and scientific consistency. 
\end{abstract}


\end{frontmatter}
\thispagestyle{fancy}
\chead{\textit{\textcopyright 2021 This manuscript version is made available under the CC-BY-NC-ND 4.0 license https://creativecommons.org/licenses/by-nc-nd/4.0/}}
%\linenumbers

\section{Introduction}\label{sec:introduction}
\begin{figure}[h]
\centering
\includegraphics[trim=10 100 10 100, clip, width=0.8\textwidth]{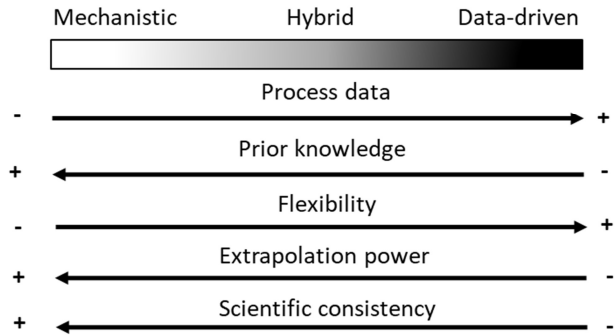}   
\caption{The range of model types from mechanistic, white-box models to data-driven, black-box models and a few of their characteristics.} 
\label{fig:grayScale}
\end{figure}

To optimally control a petroleum asset and maximize the recovery of oil and gas, it is necessary to have an adequate understanding of the behavior of the petroleum production system. This consists of the reservoir, wells, flowlines, pipelines, and separators. Commonly, a mathematical model of the flow through the production system is developed as an aid to information gathering and analysis of the system response to changes in control variables. Such a model is often referred to as a virtual flow meter (VFM) \citep{Toskey2012}. A VFM aims to continuously predict the multiphase flow rates (mixture of gas, oil, and water) at strategic locations in the asset, for instance in individual wells. The characteristics of multiphase flow represents a particular challenge to prediction. Several types of VFM models exist, ranging from mechanistic to data-driven, thus, from white-box to black-box, respectively \citep{Prada2018}. Depending on the prior knowledge about the system and the available process data, one model type can be more suitable than others, see Figure \ref{fig:grayScale}. 

\subsection{Virtual flow meter models}
Mechanistic models are based on prior knowledge about the process and utilize first-principle laws, with possible empirical closure relations, to describe the relationship between the process input, internal, and output variables \citep{Shippen2012}. Contrarily, data-driven models require no prior knowledge of the process, and rather exploit patterns in available process data to describe the input-output relationship. Therefore, data-driven models often lack scientific consistency. A model may be considered scientifically consistent if the output of the model is plausible and in line with existing scientific principles \citep{Roscher2020}. Although this concept is hard to quantify and dependent on the user's scientific knowledge, it is an important characteristic as it promotes trust in the model. As mechanistic models are derived from physical laws, their scientific consistency is high. On the other hand, assumptions and simplifications of the process physics are typically necessary for a mechanistic model to be computationally feasible and suitable for use in real-time control and optimization applications \citep{Solle2016}. Accordingly, mechanistic models often lack flexibility, which is the ability to adapt to unknown and unmodeled physical phenomena. Oppositely, due to the generic structure of data-driven models, the flexibility is high and the models may adapt to arbitrary complex physical behavior as long as this is reflected in the available data. Yet, data-driven models are data-hungry and sensitive to the quality and variability of the data. If care is not taken, overfitting of the model to data is a frequent outcome that results in poor extrapolation abilities to future process conditions \citep{Solle2016}.  

Gray-box models, or hybrid models, are a combination of mechanistic and data-driven models. The goal is to achieve a computationally feasible model that have a high flexibility and a scientifically consistent behavior. There exist numerous ways of constructing hybrid models. According to \citet{Willard2020}, gray-box models can be divided into two domains: 1) data-driven modeling to advance first-principle models, or 2) utilization of first principles to guide data-driven models. The two domains correspond to either side of the gray-scale illustrated in Figure \ref{fig:grayScale} and will be referred to as the white-to-gray and the black-to-gray approach. Taking VFM as an example, a white-to-gray model is obtained if a mechanistic model is used as a baseline whereupon data-driven models are inserted to replace assumptions or simplifications. For instance, a common approach to estimate the density of gas in a mechanistic model is with the real gas law. Instead, if this relation is described with a data-driven model, a white-to-gray VFM model is obtained. Another example is to introduce a data-driven model to capture the error between the output of the mechanistic model and corresponding measurements, see an example in \citet{Bikmukhametov2020b}. In general, the data-driven models may substitute any factors or terms in the mechanistic model. An example of a black-to-gray VFM model is if first principles are exploited to calculate additional features to be applied as input to a data-driven model. This is commonly referred to as feature engineering. A different approach is a division into natural submodels, for instance individual wells in an asset, describe each with a data-driven model and combine the output using first-principle laws. The two approaches can also be juxtaposed. For instance, both a mechanistic and a data-driven model can be developed to predict the multiphase flow rate and the model outputs combined in an ensemble model. Independent of the gray-box model type, measures should be taken to determine an appropriate degree of influence the mechanistic and data-driven part should have on the model output. In other words, there should exist a pertinent balance between learning from physics and learning from data. For instance, if the available process data are inaccurate, the mechanistic part of the model should influence the gray-box model output the most. If the process exhibits unknown behavior, the data-driven part should have the greatest impact. Desirably, the gray-box model should learn as much as possible from both physics and data. 

\subsection{Literature review}
The literature reports substantial research on mechanistic and data-driven modeling of VFMs \citep{Amin2015, Zangl2014, Ajmi2015, AlQutami2017a, AlQutami2017b, AlQutami2017c, ALQutami2018, Omrani2018, Bikmukhametov2019, Ghorbani2019}. An extensive review is found in \citet{Bikmukhametov2020a}. Some well-known commercial mechanistic VFMs are Olga, LedaFlow, FlowMananger, ValiPerformance, and Prosper. In the study by \citet{Amin2015}, it was found that all the above-mentioned commercial mechanistic VFM achieved an error less than 5\% and 10\% for the prediction of oil and gas flow rates, respectively. The noticeable series of studies on data-driven VFM by \citet{AlQutami2017a, AlQutami2017b, AlQutami2017c, ALQutami2018} achieved errors of 1.5\%, 4.2\%, and 4.7\% on the predictions of gas, oil, and water flow rates, respectively. 

Despite recent emerging tools for hybrid, gray-box modeling, such as gPROMS \citep{gPROMS}, and even a commercially available hybrid VFM: TurbulentFlux \citep{TurbulentFlux}, little literature on the performance of gray-box VFMs exist. TurbulentFlux reports an error of 4\% on multiphase flow rate predictions over two months for one of the tested wells. However, the robustness in performance for different wells is not reported. Furthermore, as no reference model is tested on the available data it is difficult to conclude whether the hybrid model performs better than alternative approaches. Nevertheless, some examples exist in the literature \citep{Xu2011, Al-Rawahi2012, Kanin2019, Bikmukhametov2020b}. Most of these studied different gray-box approaches on synthetic data, either as an experimental set up in a test rig \citep{Xu2011} or a multiphase flow loop \citep{Al-Rawahi2012}, or using lab data available online \citep{Kanin2019}. The study in \citet{Bikmukhametov2020b} investigated several hybrid VFM variants on real production data, with a large focus on the black-to-gray modeling approach. However, their results were based on process data from only one subsea well and the modeling approach could benefit from a deeper study of more petroleum wells. 

\subsection{Contributions}
This research contributes to the field of gray-box VFM modeling with an in-depth study of five white-to-gray VFM models of a petroleum production choke valve. A mechanistic and data-driven model is developed for comparison of the performance and scientific consistency. The study is a significant expansion of the work done in \citet{Hotvedt2020a, Hotvedt2020b}. The number of tested gray-box models is increased, the complexity of the model components is higher, and data from more wells are included. The VFM models are developed for 10 petroleum wells at Edvard Grieg \citep{EG}. Real, historical production data are used in the model development, thus no experimental setup or simulator is required for data acquisition. With data from 10 wells, the robustness of the modeling approaches can be investigated to a certain extent. The results in this research are in respect to the VFM application, and the generalizability to other application areas is not considered. 

\section{Production choke valve models}\label{sec:models}
\begin{figure}[ht]
\centering
\includegraphics[width=1.0\columnwidth]{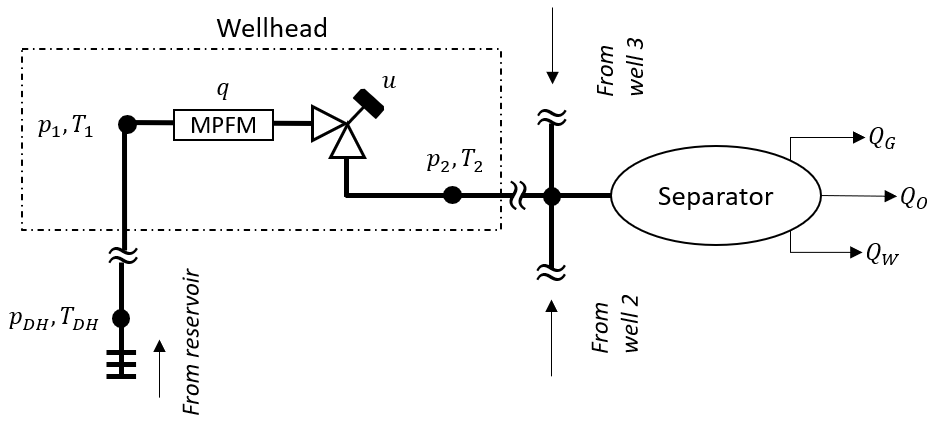}   % printed width is 8.4
\caption{Illustration of the production system, from the down-hole (DH) to the separator. The production choke valve is located in the wellhead. Typically available measurements are indicated.} 
\label{fig:choke}
\end{figure}

A production system is illustrated in Figure \ref{fig:choke}, from the down-hole, the closest measurement point to the reservoir, to the separator. The volumetric flow rate from several wells are commingled and the total production from the asset is separated into three phases, oil ($Q_{O}$), water ($Q_{W}$), and gas ($Q_{G}$), at the separator. The production choke valve is located in the wellhead of the production system. The choke is a key element in the daily control and optimization of a petroleum production system. By adjusting the choke opening, the multiphase flow rate through the production system can be controlled to maximize production while meeting operational requirements such as production capacity constraints. In this research, only the production choke is modeled. This results in lesser model complexity and avoids the utilization of down-hole sensor measurements. For assets where down-hole measurements are lacking or faulty, this is advantageous. Naturally, for assets with good down-hole measurements, the VFM can be expanded. 

To model the choke for individual wells, the following measurements are required: the choke opening ($u$), the pressures ($p$) and temperatures ($T$) located upstream (1) and downstream (2) the choke valve, and measurements of the flow rate ($q$). Measurements of the phasic flow rates in individual wells $\bm{q}=(q_O, q_G, q_W)$ can be obtained from well tests, for instance using a test separator, or multiphase flow meters (MPFMs) if these are available. 
%In some wells, multiphase flow meters (MPFM) are installed. An MPFM is a measurement device able to give continuous measurements of the phasic flow rates $q=(q_O, q_G, q_W)$. However, they may drift in time and have low accuracy in between sensor calibrations \citep{Falcone2013}. Another option to measure flow rates from individual wells is to route the production from a well to a test separator. Commonly, test separators measure the flow rates with higher accuracy, yet provide infrequent measurements.
Furthermore, the phasic fluid mass fractions are required. Ideally, these should be calculated with a different model for each new sample, for example using a simplified wellbore model as in \citet{Kittilsen2014}. Nevertheless, in this research, the mass fractions are treated as measurements, calculated using the flow rates from the MPFM in the previous measurement sample. Consequently, the utilized mass fraction will lag behind the true mass fractions. However, under the assumption of a slowly time-varying process, the mass fractions should not change significantly between each sample. 

\subsection{Mechanistic production choke model}\label{sec:models-mechanistic}
Several mechanistic models exist for the production choke, in a varying scale of complexity in space and time. Mechanistic choke models are usually developed assuming steady-state, one dimensional (lumped) flow since increasing the dimensionality of the problem requires a numerical solution of the complex Navier-Stokes equations. These equations are computationally demanding and may not be suitable for use in real-time optimization \citep{Shippen2012}. There are several well-known choke models in literature and industry \citep{Selmer-Olsen1995, Sachdeva1986, Perkins1993, AlSafran2009}. In this research, the Sachdeva model is used as the baseline model for hybridization. The Sachdeva model is one of the less complex models as it introduces many assumptions and simplifications. Expectantly, introducing data-driven elements into the mechanistic model should increase the flexibility of the model and possibly supersede some of the simplifications. The exception is distributed effects in space and time as the Sachdeva model is assumed lumped and steady-state.

The Sachdeva model is derived from the combined mass and momentum balance equations \citep[p.~107]{NodalAnalysis2015}:
\begin{equation}\label{eq:mass-momentum-balance}
\frac{dp}{ds} + \rho v \frac{dv}{ds} = 0,
\end{equation}
\begin{equation}\label{eq:mass-balance}
\dot{m} = A_1v_1\rho_1 = A_2v_2\rho_2, 
\end{equation}
in which $s$ is the position along a streamline, $\rho$ is the fluid mixture density, $v$ is the fluid mixture velocity, $\dot{m}$ is the mass flow rate, and $A$ is the area of the choke valve. Positions (1) and (2) indicate the inlet and outlet, respectively. By integrating Equation \eqref{eq:mass-momentum-balance} between location (1) and (2) and introducing several assumptions, for example:
\begin{itemize}
    \item no-slip: the gas and liquid travels through the choke with equal velocity, 
    \item incompressible liquid: liquid densities are constant along $s$ resulting in the oil and water densities being independent of the process conditions,
    \item frozen flow: no mass transfers from one phase to another across the choke resulting in constant mass fractions independent of process conditions, 
    \item adiabatic gas expansion across the choke: no mass or heat transfers between the fluid and the surroundings,
    \item thoroughly and homogeneously mixed fluid,
    \item neglect of momentum effects at (1) due to $A_2 \ll A_1$, yielding $v_2^2 \gg v_1^2$,
\end{itemize}
a model for the mass flow rate through the choke valve is obtained (see \citet{Sachdeva1986} for complete derivation):
\begin{equation}\label{eq:mass_flow_rate}
\begin{aligned}
    \dot{m} &= C_DA_2(u) \times \\
    &\sqrt{2\rho_2^2p_1\left(\frac{\kappa}{\kappa -1}\eta_{G}\left(\frac{1}{\rho_{G,1}}-\frac{p_r}{\rho_{G,2}}\right) + \left(\frac{\eta_{O}}{\rho_{O}} + \frac{\eta_{W}}{\rho_{W}}\right)(1 - p_r)\right)},
\end{aligned}
\end{equation}
\begin{equation}\label{eq:real-gas-law}
\rho_{G,1} = \frac{p_1M_G}{ZRT_1},
\end{equation}
\begin{equation}\label{eq:polytropic-gas-expansion}
\rho_{G,2} = \rho_{G,1}p_r^{\frac{1}{\kappa}},
\end{equation} 
\begin{equation}\label{eq:homogeneous-mix-density}
    \frac{1}{\rho_2} = \frac{\eta_{G}}{\rho_{G,2}} + \frac{\eta_{O}}{\rho_{O}} + \frac{\eta_{W}}{\rho_{W}}, 
\end{equation} 
\begin{equation}\label{eq:mass-fraction-summation}
   \eta_G + \eta_O + \eta_W = 1.
\end{equation}
Here $\rho_i, \eta_i, i\in \{G, O, W\}$ are the phasic densities and mass fractions, respectively, $M_G$ is the molar mass of gas, and $p_r$ is the downstream to upstream pressure ratio. The gas expansion coefficient $\kappa$ is in this article treated as a constant but is in practice a function of pressure and temperature, $\kappa = \kappa(p_1, p_2, T_1, T_2)$. The gas compressibility factor $Z$ is calculated using the correlation in \citep{Sutton1985}. The discharge coefficient $C_D$ is commonly introduced to account for modeling errors. The area of the choke is a function of the choke opening $A_2 = A_2(u)$ since the choke is adjustable. 

The model differentiates between critical and subcritical flow using
\begin{equation}\label{eq:pressure-ratio}
p_r = 
\begin{cases}
\frac{p_2}{p_1} & \frac{p_2}{p_1} \geq p_{r,c} \\
p_{r,c} & otherwise
\end{cases}
\end{equation}
In short, critical flow is a phenomenon where the mass flow rate through the choke is not increasing for decreasing downstream pressure $p_2$ and fixed upstream pressure $p_1$. A rule of thumb for the critical flow boundary $p_{r,c}$ for multiphase flow with a mixture of gas, oil, and water is $p_{r,c} \approx 0.6$ \citep{NodalAnalysis2015}. 
The volumetric flow rate may be obtained using the mass flow rate and the mixture density in standard conditions (SC), typically 1 atm and 15$^{\circ}C$ \citep{ISO13443}. In this research, the model output is the oil volumetric flow rate:
\begin{equation}\label{eq:volumetric-flow-rate-phase}
    q_{O} = \frac{\eta_{O}\dot{m}}{\rho_{O,SC}},
\end{equation}

Mathematically, the mechanistic model (MM) in \eqref{eq:mass_flow_rate}-\eqref{eq:volumetric-flow-rate-phase} is described with the generic function $f$ that predicts the oil volumetric flow rate for the input measurements $\bm{x}$ and the set of model parameters $\bm{\phi}_{MM}$:
\begin{equation}
    \hat{y}_{MM} = q_{O,MM} = f(\bm{x}; \bm{\phi}_{MM}) \in \reals,
\end{equation}
\begin{equation}\label{eq:x}
    \bm{x} = (p_1, p_2, T_1, T_2, u, \eta_{G}, \eta_{O}) \in \reals^7,
\end{equation}
\begin{equation}
    \bm{\phi}_{MM} = (\rho_{O}, \rho_{W}, \kappa, M_G, p_{r,c}, C_D) \in \reals^6.
\end{equation}
The $\bm{\phi}_{MM}$ are components in the model which are considered constant due to certain assumptions or simplifications. For instance, as described above, the oil and water densities are constant parameters due to the assumption of incompressible liquid. 

\subsection{Hybridization of the mechanistic model}\label{sec:models-hybridization}
To hybridize the MM, any of the factors or terms in \eqref{eq:mass_flow_rate}-\eqref{eq:volumetric-flow-rate-phase} can be substituted with a data-driven model (DM). Approaching the hybridization from a physical point of view, some of the mechanistic model assumptions or simplifications can be imprecise, yielding an erroneous physical behavior. For instance, in low temperature and high-pressure conditions, the real gas law relation in \eqref{eq:real-gas-law} may be inaccurate. Instead of using a different, and possibly more complex, mechanistic relation such as van der Waals equation of state, the hybrid model utilizes a DM to substitute the real gas law. Presumably, by learning the gas density relation from patterns in the measurements only, a relation that is suitable for the process and adaptable to the current conditions is obtained. Taking another example, the adiabatic gas expansion equation in \eqref{eq:polytropic-gas-expansion} assumes that no heat or mass transfer occurs between the system and surroundings, yet, in practice, both exist. If the available measurements reflect these physical phenomena, a DM substituting \eqref{eq:polytropic-gas-expansion} should, to some extent, be able to implicitly capture the effect of, for instance, heat transfer on the flow rate, even without measurements of the ambient temperature. Similarly, most of the assumptions listed above may be replaced with a data-driven model to account for erroneous physics. Consequently, the model should be more generic and suitable for utilization in a larger range of process conditions. Nevertheless, data-driven models are generally only valid in the domain of the data they have been exposed to. Hence, if the system is exposed to previously unseen process conditions, the hybrid models will likely have to be retrained or recalibrated to adapt to the new data. 

There is an abundant number of hybridization options of the mechanistic model. Therefore, only a few of the simplifications and assumptions of the baseline model are investigated. Further, numerous combinations of these simplifications are viable, and for simplicity, only one simplification is considered at the time. Thereby, five hybrid model (HM) variants are developed, each addressing and substituting one of the following simplifications with a DM:

\begin{enumerate}
    \item The area function, $A_2(u)$
    \item The upstream gas density function, replacing  \eqref{eq:real-gas-law}
    \item The adiabatic gas expansion function, replacing \eqref{eq:polytropic-gas-expansion}. 
    \item The homogeneous mixture density function, replacing \eqref{eq:homogeneous-mix-density}.
    \item An additive error model to capture structural errors of the MM
\end{enumerate}

Mathematically, the inserted DM is defined by
\begin{equation}
    \hat{y}_{DM} = g(\bm{x}_{DM}; \bm{\phi}_{DM}) \in \reals,
\end{equation}
where $\bm{x}_{DM}\subseteq \bm{x}$ depends on the HM variant, and $\bm{\phi}_{DM}$ are a set of nonphysical parameters defining the structure of the DM. For the interested reader, if there exist measurements of what the DM represents, for example, density measurements, these may be incorporated into the model development by the means of prior parameter specification. 

The HM is defined as a combination of the MM and DM by:
\begin{equation}
\begin{aligned}
    \hat{y}_{HM} &= q_{O,HM} = h(\bm{x}_{HM}; \bm{\phi}_{HM}) \in \reals,
\end{aligned}
\end{equation}
where $\bm{x}_{HM}\subseteq\bm{x}$ and the hybrid model parameters $\bm{\phi}_{HM}$ is all of $\bm{\phi}_{DM}$ but not necessarily all of $\bm{\phi}_{MM}$ since some are redundant when introducing the DM in the MM. For instance, replacing \eqref{eq:real-gas-law} with a DM, the parameter $M_G$ is no longer needed in the equations. 

The five HMs may be illustrated with the following figures, variant 1-4 in Figure \ref{fig:HM1-4-illustration}, here $\bm{\phi}'_{MM} \subseteq \bm{\phi}_{MM}$, and HM variant 5 in Figure \ref{fig:HM5-illustration}. It should be noted that the framework used to develop the gray-box models are not restricted to the variants in Figure \ref{fig:HM5-illustration}. For instance, only small changes to the model are necessary to  implement black-to-gray VFM models. 

\begin{figure}[ht]
\centering
\subfloat[HM 1-4]{
%left, bottom, right, top
\includegraphics[trim=0 100 0 100, clip, width=0.53\textwidth]{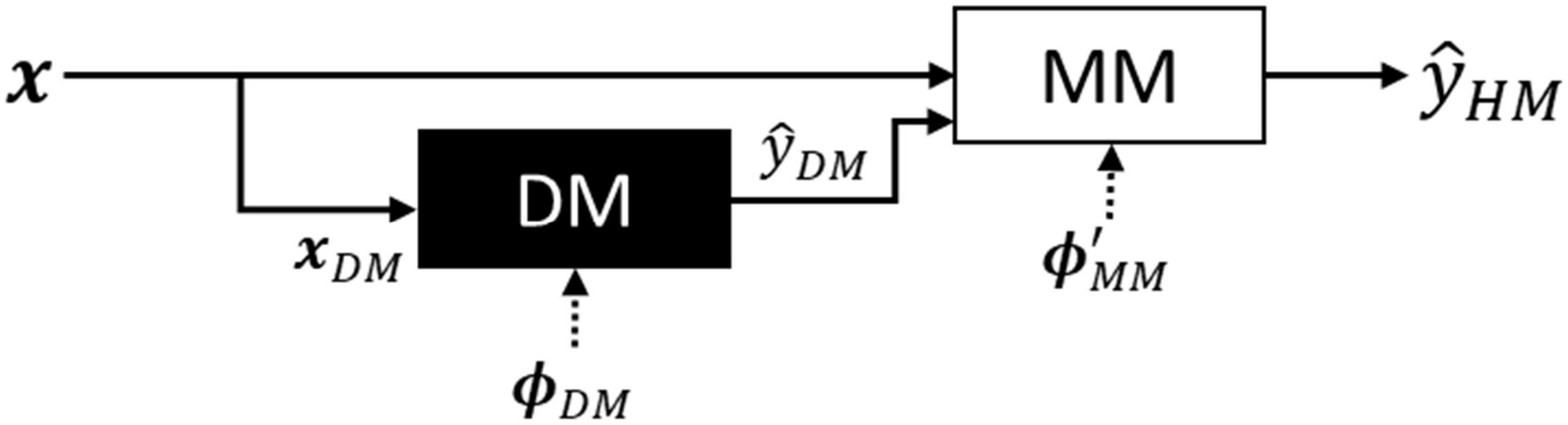}
\label{fig:HM1-4-illustration}
}
\hfill 
\subfloat[HM 5]{
\includegraphics[width=0.43\textwidth]{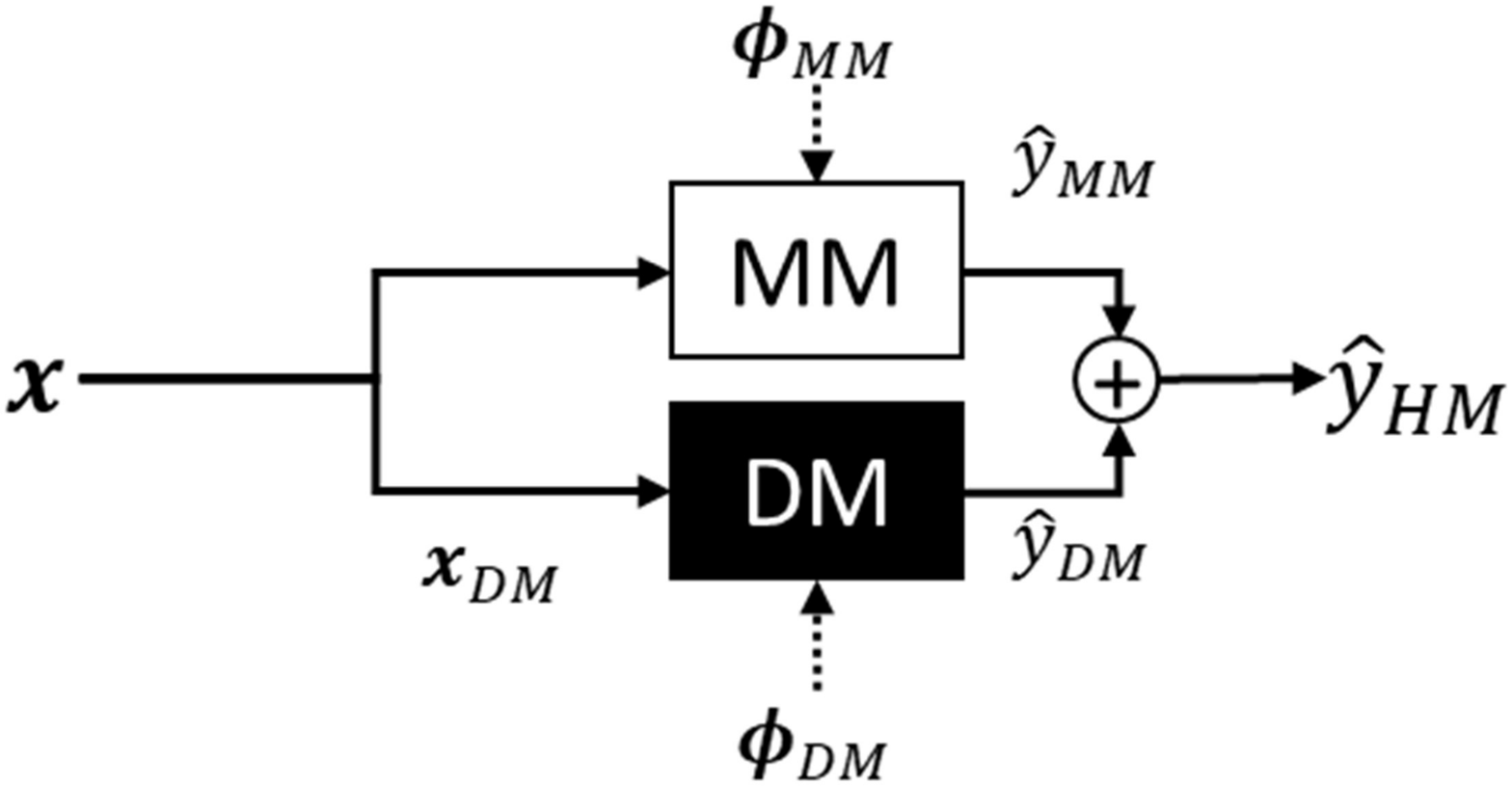}
\label{fig:HM5-illustration}
}
\caption{Illustration of the five hybrid model variants. (a) Hybrid model variant 1-4, (b) Hybrid model variant 5: additive error model.}
\label{fig:HMtypes}
\end{figure}

The applied data-driven model for all the hybrid model variants is a fully connected, feed-forward neural network. Naturally, other data-driven methods may be applied such as regression trees or support vector machines. Nevertheless, as mentioned in Section \ref{sec:introduction}, neural networks are flexible and can adapt to arbitrarily complex patterns in data. Furthermore, the neural network is easily integrated into a model development framework where the model parameters are found with maximum a posteriori estimation and stochastic gradient-based optimization. This will be introduced in Section \ref{sec:parameter-estimation}. In short, a feed-forward neural network is a collection of L layers, represented with the following equations:
\begin{equation}\label{eq:nn-layer}
\begin{aligned}
    \text{Input}   \quad z_0 &= \bm{x}_{DM}\\ 
    \text{Hidden layer(s)} \quad \bm{z}_i &= a_i(\bm{W}_i\bm{z}_{i-1} + b_i), \quad i\in\{1,..,L-1\} \\
    \text{Output layer} \quad z_L &= \bm{W}_L\bm{z}_{L-1} + b_L \\ 
\end{aligned}
\end{equation}
At each layer, the inputs are transformed with a linearly affine function with weight matrix $\bm{W}_i$ and bias $b_i$ and sent through an activation function $a$. The rectified linear unit activation function has been used, which is the elementwise maximum operator $\text{ReLU}(\bm{z}_i) = \max\{0, \bm{z}_i\}$. This results in the neural network being a set of piecewise linear equations. The nonphysical parameters of the network are the collection of weights and biases on all layers $\bm{\phi}_{DM} = \{(\bm{W}_1, b_1),\ldots (\bm{W}_L, b_L)\}$.  

\section{Parameter estimation of hybrid models}\label{sec:parameter-estimation}
Regardless of the location of the model on the gray-scale in Figure \ref{fig:grayScale}, the uncertain model parameters should be estimated from data. For a fully mechanistic model, good prior values on the parameters often exist and parameter estimation is not a requirement, although usually a necessity, for high accuracy model predictions. For a fully data-driven model, parameters are initialized randomly and parameter estimation is a requirement. Thus, the latter argumentation applies to hybrid models. Parameter estimation is also referred to as model training.

\subsection{Maximum a posteriori estimation}\label{sec:map}
Consider a dataset $\mathcal{D} = \{\bm{x}_i, y_i\}_{i=1}^n$ with $n$ measurements of the process explanatory variables $\bm{x}_i = (x_{i,1}, \ldots x_{i,d}) \in \mathbb{R}^d$, and target variable $y_i \in \mathbb{R}$. Assume the process to be described by the following measurement model
\begin{equation}\label{eq:general-model}
    y_i = h(\bm{x}_i; \bm{\phi}) + \epsilon_i, \quad \epsilon_i \sim \mathcal{N}(0,\sigma_{\epsilon,i}^2) \quad i \in \{1,..,n\},
\end{equation}
where $\hat{y}_i = h(\bm{x}_i; \bm{\phi})$ are the model predictions of the target variable, with model parameters $\bm{\phi}\in\mathbb{R}^m$ and normally distributed measurement $\epsilon_i$ with zero mean and variance $\sigma_{\epsilon,i}^2$. Observe that this measurement model can incorporate output measurement from different devices, if available, by changing $\sigma_{\epsilon, i}^2$ appropriately for measurement $i$. Even synthetic data generated with mechanistic simulators may be included in this approach.  

In parameter estimation problems, the parameters $\bm{\phi}$ of the model $h$ will be inferred using the available data $\mathcal{D}$. This can be done using Bayesian inference where the prior parameter distribution $p(\bm{\phi})$ is updated to a posterior parameter distribution:
\begin{equation}\label{eq:Bayes}
    p(\bm{\phi} \condbar \mathcal{D}) = \frac{p( \mathcal{D} \condbar \bm{\phi})p(\bm{\phi})}{p(\mathcal{D})}.
\end{equation}
Equation \eqref{eq:Bayes} includes intractable integrals \citep{Blei2017} and approximation techniques are commonly required for a numerical solution. In this research, maximum a posteriori (MAP) estimation is applied. 

In MAP estimation, only the mode of the posterior distribution is considered and the parameters are found with the following optimization problem:
\begin{equation}
\begin{aligned}
    \bm{\phi}^{\star}_{MAP} &= \arg \max_{\bm{\phi}} p(\bm{\phi} \condbar \mathcal{D}) = \arg \max_{\bm{\phi}} \Big[\log p( \mathcal{D} \condbar \bm{\phi}) + \log p(\bm{\phi})\Big], 
\end{aligned}
\end{equation}
where $\log p( \mathcal{D} \condbar \bm{\phi})$ is called the loglikelihood of the model. By further assuming normally distributed parameter priors $\phi_i \sim \mathcal{N}(\mu_i, \sigma_i^2), i\in\{1,..,m\}$ the following optimization problem may be derived \citep{Bishop2006}:
\begin{equation}\label{eq:MAP}
\begin{aligned}
 \bm{\phi}_{MAP}^* &= \arg \min_{\bm{\phi}} \left[\sum_{i=1}^n \frac{1}{\sigma_{\epsilon, i}^2}\left(y_i - f(\bm{x}_i;\bm{\phi})\right)^2 + \sum_{i=1}^{m} \frac{1}{\sigma_{i}^2}\left(\phi_i - \mu_i\right)^2\right].
\end{aligned}
\end{equation}
In short, MAP estimation is a trade-off between minimizing the error between target variable predictions and measurements and minimizing parameter deviation from the prior mean $\mu$. By setting a constant noise level $\sigma_{\epsilon}^2 = const.$, the MAP estimation is equal to maximum likelihood estimation (MLE) with $\ell_2$-regularization, a common approach in the data-driven modeling domain \citep{Goodfellow2016}. The variance of the parameters and measurement noise determine the degree of regularization. In \citet{Hotvedt2020b}, it was shown that MAP estimation is necessary for a hybrid model to obtain plausible and physically consistent values of the physical model parameters after estimation. Further, regularization must be used to avoid overfitting of the model and ensure adequate generalization performance \citep{Goodfellow2016}. 

A different perspective of the MAP estimation problem is that it balances learning from physics and learning from data. With softer regularization, achieved by setting flat, noninformative prior parameter distributions $\sigma_{i}\to\infty$, the data will have a large influence on the estimation outcome. This is because the regularization terms are down-weighted in optimization. The same effect is achieved with a small noise variance, implying that the measurements are accurate. With harder regularization, the opposite effect is achieved where the physics, in this case, the parameter priors, will have a higher influence on the estimation outcome and the adaption to data down-weighted in the optimization. 

For the HMs in Section \ref{sec:models}, the MAP objective function is divided into three terms, the MLE and two parameter regularization terms, one each for the physical and nonphysical parameters. In this research, only MPFM measurements are used and thus:
\begin{equation}\label{eq:map-hybrid}
\begin{aligned}
\bm{\phi}_{MAP}^* & =  \arg \min_{\bm{\phi}} \sum_{i=1}^n \Big(y_i - h(\bm{x}_{i, HM};\bm{\phi}_{HM})\Big)^2 \\
&+ \sigma_{\epsilon}^2 \Bigg[\sum_{i=1}^{m_1} \left(\frac{\phi_{i,MM} - \mu_{i,MM}}{\sigma_{i, MM}}\right)^2 + \sum_{i=1}^{m_2}\left( \frac{\phi_{i,DM} - \mu_{i, DM}}{\sigma_{i,DM}}\right)^2\Bigg].
\end{aligned}
\end{equation}
Here $m_1$ and $m_2$ is the number of physical and nonphysical parameters, respectively. 
%For the interested reader, the objective function in Equation \eqref{eq:map-hybrid} is valid for other types of gray-box models as well, for instance, ensemble models. In such a case, the MLE may be divided into two, one each for the output of a mechanistic and a data-driven model, and the terms appropriately weighted according to the desired influence on the estimation outcome. 

\subsection{Priors on the physical parameters}\label{sec:parameter-estimation-par-phys-prior}
For the physical model parameters, good prior values of the mean $\mu_{i, MM}$ often exist. For instance, for freshwater density $\mu_{\rho_w} \approx 1000$ kg/m$^3$. The parameter variances may be set to reflect the uncertainty in the prior mean value. If the assumption of normally distributed parameters is exploited, the variance may be approximated using the absolute maximum and minimum values of the parameters and calculating the $6\sigma$ band of the distribution, 
\begin{equation}
    \sigma_{i, MM} = \frac{\max{(\phi_{i,MM})} - \min{(\phi_{i, MM})}}{6},
\end{equation}
for which the probability of obtaining values outside the band is $\approx 0.03\%$. For harder regularization of a specific parameter, the variance may be decreased, resulting in a sharper distribution, and the opposite for softer regularization. 

\subsection{Priors on the nonphysical parameters}\label{sec:parameter-estimation-par-nn-prior}
Finding priors for the nonphysical parameters in the model is not trivial. However, He-initialization is recommended for neural networks with ReLU as activation function \citep{He2015}. With He-initialization, each element in the weight matrix on each layer $\bm{W}_i, i\in L$ (see Section \ref{sec:models}) is initialized from a normal distribution with mean and variance  
\begin{equation}\label{eq:par-nn-prior}
\begin{aligned}
    \mu_{DM} = 0, \qquad \sigma_{DM}^2 = \left(\sqrt{\frac{2}{m_{l,i}}}\right)^2), \quad i \in {2, .. L},
\end{aligned}
\end{equation}
where $m_{l,i}$ are the number of inputs on layer $i$. On the first layer, no activation function is applied to the inputs and  $\sigma_{DM}^2 = \left(\sqrt{1/m_{l,1}}\right)^2)$. 

On the other hand, for the hybrid models where the neural network represents a mechanistic relation, more informative priors for the nonphysical parameters are found by pretraining the network on synthetic data generated with the mechanistic relation in question. The obtained values of the weights and biases of the pretrained network are used as the $\bm{\mu}_{DM}$ when training the final model. However, the the network is trained on synthetic data only and it assumed that the updated prior parameter means are just as uncertain as before. Therefore, the parameter variances in \eqref{eq:par-nn-prior} are utilized. If real measurements of the variable existed, such as density measurements, these could be used in the pretraining. 

\subsection{Priors on the measurement noise}\label{sec:parameter-estimation-par-noise-prior}
In an industrial setting, a common measure of the error of a measurement device is the mean absolute percentage error (MAPE), comparing the measured signal to a known reference value $y_{ref}$. Following the derivation in \citet{Grimstad2021}, the MAPE may be translated into the variance of the measurement noise with
\begin{equation}
    \sigma_{\epsilon}^2 = \left(\sqrt{\frac{\pi}{2}}\alpha |y_{ref}|\right)^2, 
\end{equation}
where $\alpha$ is the MAPE, for instance $\alpha = 0.1$ for 10\% MAPE. In this study, the reference value is not known and the variance of the measurement noise is approximated by using the available data. Because the MAP estimation in \eqref{eq:map-hybrid} assumes a constant noise level $\sigma_{\epsilon}^2 = const.$, the mean value of the measured target variable in the training data is used as the reference value, $y_{ref} = 1/n \sum_{i=1}^n y_i$. As mentioned in Section \ref{sec:map}, in practice the $\sigma_{\epsilon}^2$ may be adjusted to influence the degree of regularization on the parameters. 

\section{Case study}\label{sec:case-study}
The case study develops the five listed white-to-gray VFM models in Section \ref{sec:models} for 10 petroleum wells on Edvard Grieg \citep{EG}. Edvard Grieg is an asset on the Norwegian Continental Shelf and consists of under-saturated oil without a gas cap. The asset is relatively new where production commenced in 2015. The wells, hereafter referred to as W01-W10, are well-instrumented with available measurements of the explanatory variables defined in \eqref{eq:x}. An MPFM located in the wellhead of each well provides measurements of the volumetric flow rate. The models are trained with MAP estimation introduced in Section \ref{sec:parameter-estimation} using real, historical production data from the 10 wells. The number of data samples per well is unequal and spans approximately 1.5-4 years. No additional experimental or synthetic data are considered. For comparison, the Sachdeva model in Section \ref{sec:models}, and a fully connected feed-forward neural network, are implemented. Two aspects of the models are investigated. First, the predictive performance in terms of accuracy is analyzed in Section \ref{sec:case-study-predictive-performance}. Thereafter, the scientific consistency is examined in Section \ref{sec:case-study-scientific-consistency}. Considerations for improvements in future work are discussed in Section \ref{sec:case-study-improvements}.  

The datasets for each well are preprocessed in two steps. First, the processing technology in \citet{Grimstad2016}, is utilized to generate a compressed dataset of steady-state operating points suitable for steady-state modeling. Secondly, a set of filters are applied to remove data samples that likely originate from erroneous sensor data, such as negative pressures or choke openings. The dataset is split into training and test set according to time to mimic an industrial setting where the developed models are used to predict the future responses of the process. The test set consists of the three latest months of the data samples. The regularization method early stopping \citep{Goodfellow2016} is utilized to train the models. This algorithm monitors the error on a validation dataset during model training to find the appropriate number of loops through the training data, called epochs, to train the model without overfitting. The validation data is 20\% of the training data, extracted in randomly chosen chunks, each chunk representing data samples from two chronological weeks. Due to the stochasticity of the training algorithm, the early stopping algorithm is run several times, and the average number of epochs is used to train the final model. The optimizer Adam \citep{adam} is applied with mini-batches, and the learning rate is $\alpha = 10^{-4}$.

An overview of the seven implemented models is found in Table \ref{tb:EG-model-types}. The table illustrates which mechanistic model parameters $\bm{\phi}_{MM}' \subseteq \bm{\phi}_{MM}$, are present in the model, which factor or term is replaced by a neural network $g$, and which measurements $\bm{x}_{DM}$ are used as input to the data-driven element. For short, the hybrid models are named $HM(\star)$, where $\star$ is the factor or term the neural network substitutes. The fully mechanistic and the fully data-driven model are referred to as the MM and the DM respectively. 
\begin{table}[ht]
\footnotesize
\caption{An overview of the developed models of the production choke valve: five hybrid, one fully mechanistic, and one fully data-driven model.}
\begin{tabularx}{\textwidth}{l*3{>{\raggedright\arraybackslash}X}}
\toprule
VFM model &  $\bm{\phi}_{MM}'$ & $g(\bm{x}_{DM}; \bm{\phi}_{DM})$ & $\bm{x}_{DM}$ \\
\midrule 
MM & $\rho_O, \rho_W, \kappa, M_G, p_{r,c}, C_D$  & n.a. & n.a.\\
HM($A_2$) & $\rho_O, \rho_W, \kappa, M_G, p_{r,c}$ & Area function & $u$\\
HM($\rho_{G,1}$) & $\rho_O, \rho_W, \kappa, p_{r,c}, C_D$ & Upstream gas density & $p_1,T_1$\\
HM($\rho_{G,2}$) & $\rho_O, \rho_W, M_G, p_{r,c}, C_D$ & Gas expansion & $p_1,  p_2, T_1, T_2$\\
HM($\rho$) & $\rho_O, \rho_W, \kappa, M_G, p_{r,c}, C_D$ & Mixture density & $p_1, p_2, T_1, T_2, \eta_G, \eta_O$\\
HM($\varepsilon$) & $\rho_O, \rho_W, \kappa, M_G, p_{r,c}, C_D$ & Additive error & $p_1, p_2, T_1, T_2, \eta_G, \eta_O$\\
DM & n.a. & Oil flow rate & $p_1, p_2, T_1, T_2, u, \eta_G, \eta_O$\\ 
\bottomrule \noalign{\smallskip}
\end{tabularx}
\label{tb:EG-model-types}
\end{table}
For all neural networks, the network depth and width are set to $3\times100$. The size may be excessive for some of the models. Nonetheless, following recommendations from \citep{Bengio2012} the size can be set arbitrarily large as long as regularization is employed to prevent overfitting. For the HM($A_2$), HM($\rho_{G,1}$), HM($\rho_{G,2}$), and HM($\rho$), the neural networks are pretrained with synthetic data before utilized in the final model. For each of the final 70 choke models (for 10 wells and 7 model types), the parameters are initialized using the prior parameter distributions described in Section \ref{sec:parameter-estimation-par-phys-prior} and \ref{sec:parameter-estimation-par-nn-prior}. The variance of the measurement noise $\sigma_{\epsilon}^2$ is calculated assuming a MAPE of 10\% and following the procedure in Section \ref{sec:parameter-estimation-par-noise-prior}. A trick is utilized to enforce the positivity of the physical model parameters. A temporary parameter $S$ is learned instead of the real parameter $\phi$, and the transformation 
\begin{equation}
    \phi_i = \exp{(S_{{\phi}_i} + \zeta)}, \quad \text{for } i=1,..,m,
\end{equation}
is used to obtain the real parameter value. Here $\zeta$ is a small constant to avoid vanishing gradients in the optimization problem. 

\subsection{Predictive performance}\label{sec:case-study-predictive-performance}
\begin{figure}[h!]
\centering
\includegraphics[width=\textwidth]{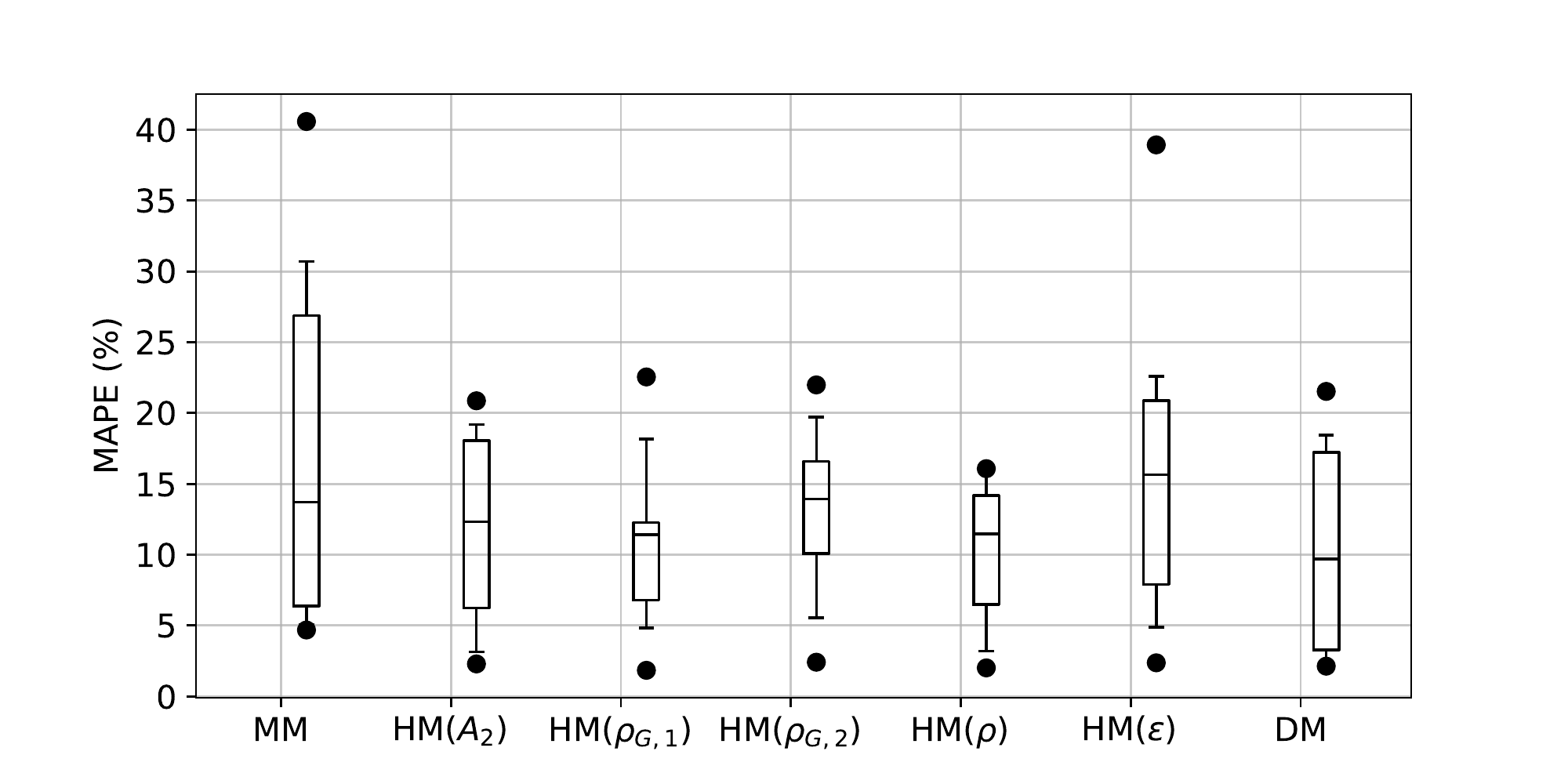}   
\caption{Box plot of the mean absolute percentage error for each model across all wells. The horizontal line in the box is the median performance.} 
\label{fig:eg-box-plot-all-wells}
\end{figure}

In Figure \ref{fig:eg-box-plot-all-wells}, the mean absolute percentage error (MAPE) is calculated for each choke model and illustrated in a box plot comparing the different model types. Table \ref{tb:EG-error} shows a detailed view of the MAPEs for the individual choke models. For the interested reader, the predicted volumetric flow rates are illustrated together with the measured flow rate (downscaled) in \ref{app:eg-results}, Figure \ref{fig:app-eg-flow}. 

There are several interesting observations to make. Firstly, the median errors are large for all model types and not at the level with the reported errors in literature, see Section \ref{sec:introduction}. Figure \ref{fig:eg-box-plot-all-wells} shows that the DM is the only model achieving a median MAPE below 10\%, though barely with 9.4\%. Secondly, the results indicate that moving the model on the gray-scale from white to gray does improve the average performance significantly, see Table \ref{tb:EG-error}. The MM achieves an error of 17.2\% against 10.3\% for the best HM. However, comparing the HMs to the DM with an error of 10.4\%, there is only a small improvement. Thirdly, large variations in performance for the different choke models are observed in Table \ref{tb:EG-error}. For instance, for W01, all the model types perform excellently and are on the level with the reported errors in the literature (less than 4\% MAPE). Yet, for W02, the performance is unsatisfactory for all model types. The large differences in performance may also be observed by looking at the cumulative deviation plots in Appendix \ref{app:eg-results}, Figure \ref{fig:eg-cum-perf-all-wells}. This plot shows the percentage of test points that fall within a certain percentage deviation from the true value \citep{Corneliussen2005}.

\begin{table}[ht]
\footnotesize
\caption{Mean absolute percentage error for the individual choke models. The best performing choke model is highlighted in bold.}
\begin{tabularx}{\textwidth}{l*7{>{\raggedleft\arraybackslash}X}}
\toprule
& MM & HM($A_2$) & HM($\rho_{G,1}$) & HM($\rho_{G,2}$) & HM($\rho$) & HM($\varepsilon$) & DM \\ \midrule
W01  & $4.7$ & $3.1$ & $\bm{1.8}$ & $2.4$ & $2.0$ & $4.9$& $2.4$  \\
W02  & $28.9$ & $16.7$ & $\bm{11.2}$ & $19.7$& $14.4$ & $18.0$ & $17.7$ \\
W03  & $20.8$ & $\bm{9.3}$ & $18.2$ & $16.7$ & $16.1$ & $22.6$ & $15.7$  \\
W04  & $\bm{5.7}$ & $18.5$ & $11.6$ & $15.5$& $13.7$ & $13.6$ & $18.5$ \\
W05  & $8.3$ & $11.9$ & $12.5$ & $22.0$ & $16.1$ & $17.4$ & $\bm{3.6}$ \\
W06 & $40.6$ & $20.9$ & $6.5$ & $9.9$& $9.3$& $38.9$ & $\bm{3.7}$ \\
W07 & $30.7$ & $2.3$ & $4.8$ & $5.5$ & $5.7$ & $6.0$& $\bm{2.1}$ \\
W08 & $5.1$ & $5.2$ & $7.7$& $10.7$& $3.2$ & $\bm{2.4}$ & $3.2$ \\
W09 & $12.7$ & $12.8$ & $11.6$ & $12.3$& $\bm{8.8}$ & $13.9$ & $21.5$ \\
W10 & $14.7$ & $19.2$ & $22.5$ & $16.3$ & $\bm{13.6}$& $21.9$ & $15.8$ \\ 
\midrule 
Across wells & $17.2$ & $12.0$ & $10.9$ & $13.1$ & $10.3$& $16.0$ & $10.4$ \\
\bottomrule \noalign{\smallskip}
\end{tabularx}
\label{tb:EG-error}
\end{table}

There are several factors that may cause the observed prediction accuracy of the different models. Three of these will be discussed in the following. Section \ref{sec:case-study-predictive-performance-model-simpifications} will focus on the impact model simplifications may have on the accuracy, Section \ref{sec:case-study-predictive-performance-balancing} will elaborate on the task of balancing learning from physics and learning from data, and Section \ref{sec:case-study-predictive-performance-data} discusses the likely influence of available data.

\subsubsection{The possible impact of model simplifications}\label{sec:case-study-predictive-performance-model-simpifications}
First of all, it must be kept in mind that only the production choke valve is modeled, and any effects of the remaining production system on the multiphase flow, such as the wellbore, are disregarded. It is believed that the average predictive performance would improve by modeling a larger part of the production system. Second of all, several assumptions and simplifications are introduced in the baseline mechanistic choke model. Dependent on process conditions, flow regimes, and fluid composition, these may be appropriate to describe the physical behavior of the flow through the choke in some wells but imprecise in others. For instance, observe how the HM($A_2$) for W03 has a much better performance than any of the other model types. This may indicate that the mechanistic area function is poorly calibrated for this well in the other model types. For W01, HM($\rho_{G,1}$) has the best performance and may suggest that the assumption of the real gas law is inadequate. Naturally, these are only indications and the results could benefit from a deeper analysis of the suitability of different hybrid models in different cases. 

\subsubsection{The nontrivial task of balancing learning from physics and data}\label{sec:case-study-predictive-performance-balancing}
With adequate design and training, the HMs were expected to exploit both physics and data to their full extent and thereby perform better than non-hybrid models. Certainly, on a well level, six wells perform better with an HM. However, seen from Table \ref{tb:EG-error}, wells W04-W07 perform better with either a mechanistic or a data-driven model. This may cast light upon the nontrivial task of balancing learning from physics and data. The HM may be too simplistic, and consequently, not flexible enough to capture complex physical behavior. Likewise, the data-driven elements may be erroneously influenced by the data. Hence, an appropriate approach to control the influence of the mechanistic and data-driven component is yet to be discovered, at least for the white-to-gray hybrid model types investigated in this research.

\subsubsection{The influence of the available data}\label{sec:case-study-predictive-performance-data}
As neural networks have the power to adapt to arbitrarily complex patterns in the data, the large MAPEs seen for many of the DMs may indicate that the quality of the available data is inadequate. Real, historical production data are used in both model training and testing. It is not uncommon that production data are noisy and biased, which complicates the modeling process and may yield an unfair indication of predictive performance for some models. Naturally, different model types or estimation techniques exist which to a greater extent exploits uncertainty in the model parameters and measurements. On the other hand, such methods require specifications of uncertainty that are not easily available, and the resulting models are usually of higher complexity. Further, it is believed that the large error for several of the choke models is mainly caused by the datasets originating from the underlying, nonstationary process. In time with the reservoir being depleted, the pressure in the down-hole will decrease. If the goal is to maintain a steady production rate, the operators must increase the choke opening. Extracting the test dataset chronologically may therefore result in a set of process conditions that are substantially different from the conditions seen in the training dataset. If so, a steady-state model like the baseline mechanistic model or a standard neural network will not be able to capture the slowly varying, underlying changes. 

\begin{figure}[ht!]
\centering
\includegraphics[width=\textwidth]{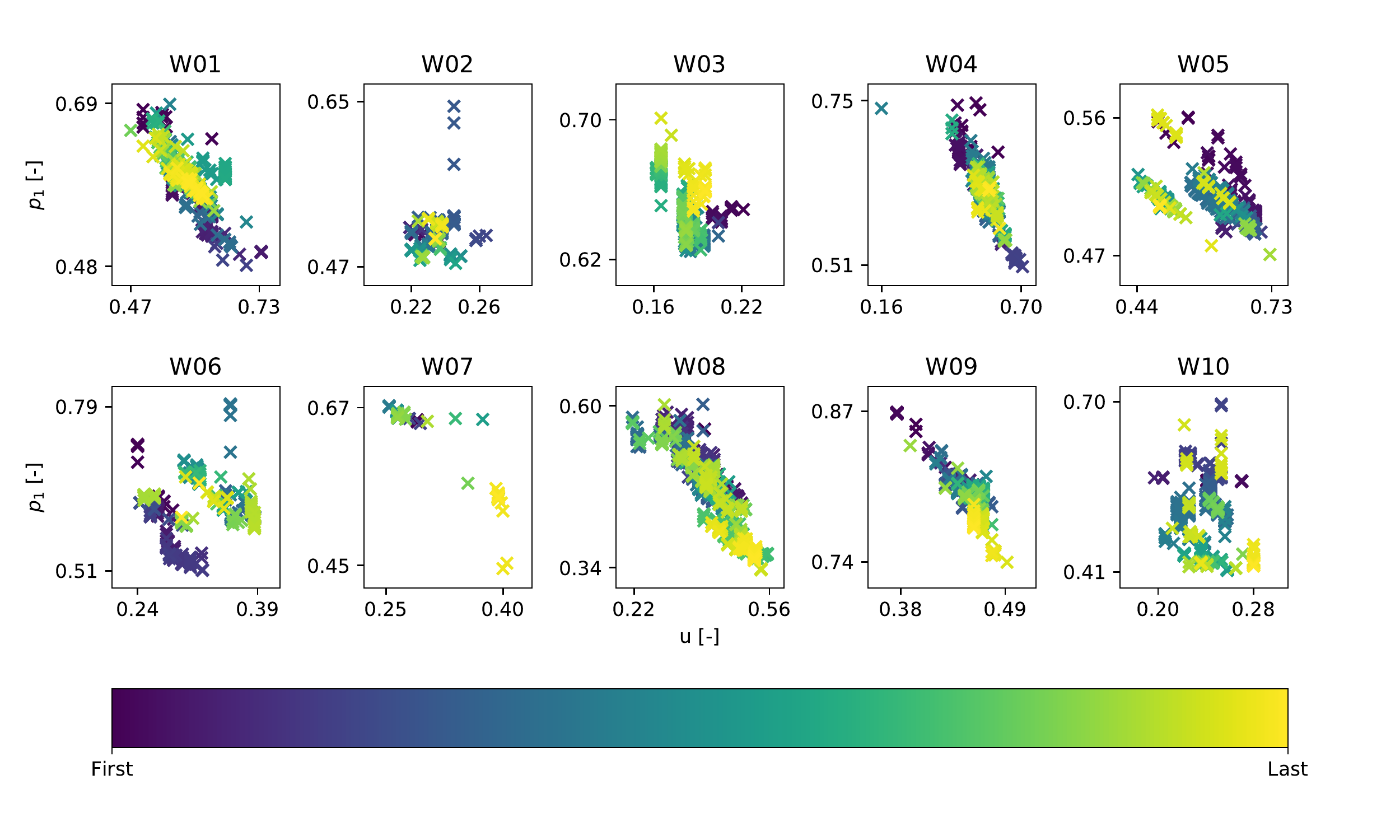}
\caption{Upstream pressure versus choke opening in time for approximately the same volumetric oil flow rate. Dark colors are the earliest time samples, whereas the light colors are the latest and are included in the test set.}
\label{fig:dataset-non-stationary}
\end{figure}
\begin{figure}[!ht]
\centering
\includegraphics[width=0.9\textwidth]{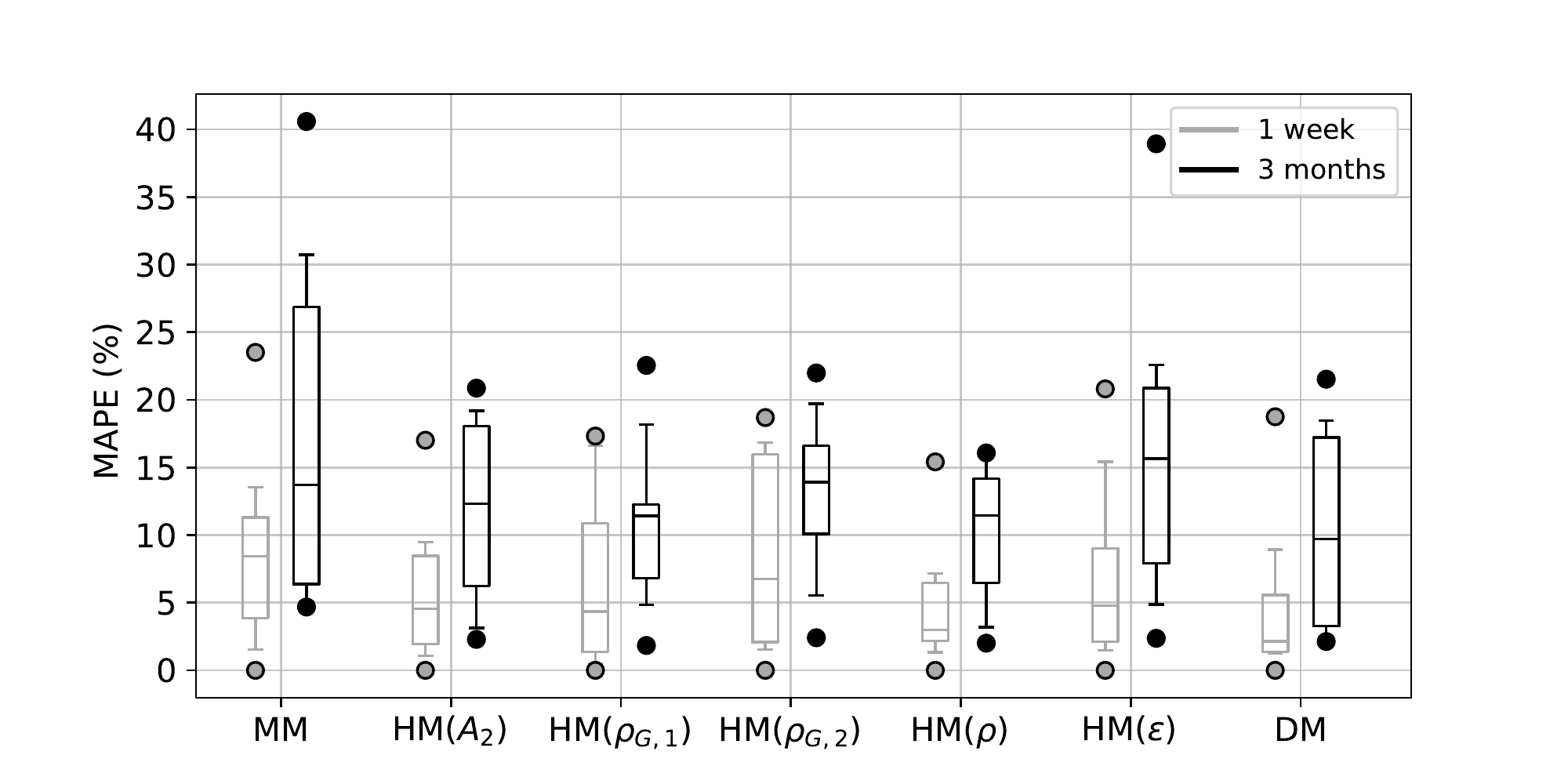}
\caption{Box plot comparing the mean absolute percentage error for each model across wells using three months of data in black (right box) and one week of data in gray (left box).}
\label{fig:box-plot-3m1w}
\end{figure}

Figure \ref{fig:dataset-non-stationary} illustrates this issue. Shown is the upstream pressure $p_1$ versus the choke opening $u$ for approximately the same oil volumetric flow rate. The coloring indicates time, the lightest colors are the latest time samples. Notice that for some wells (for example W05, W06, W07), the coloring is grouped, indicating that in time, different process conditions are required to maintain the volumetric oil flow rate. Naturally, the flow rate will also depend on other variables such as the mass fractions. Nevertheless, in a nonstationary situation, using three months of test data and assuming the model parameters to be constant and representative for the physical behavior during three months may be inappropriate. It may also discredit the high accuracy prediction potential of the models. Using the developed models to predict the process response only one week ahead greatly increases the accuracy, see the comparison of three months prediction against one-week predictions in Figure \ref{fig:box-plot-3m1w}.

\subsection{Scientific consistency}\label{sec:case-study-scientific-consistency}
\begin{figure}[!ht]
\centering
\subfloat[HM($A_2$)]{\includegraphics[width=0.48\textwidth]{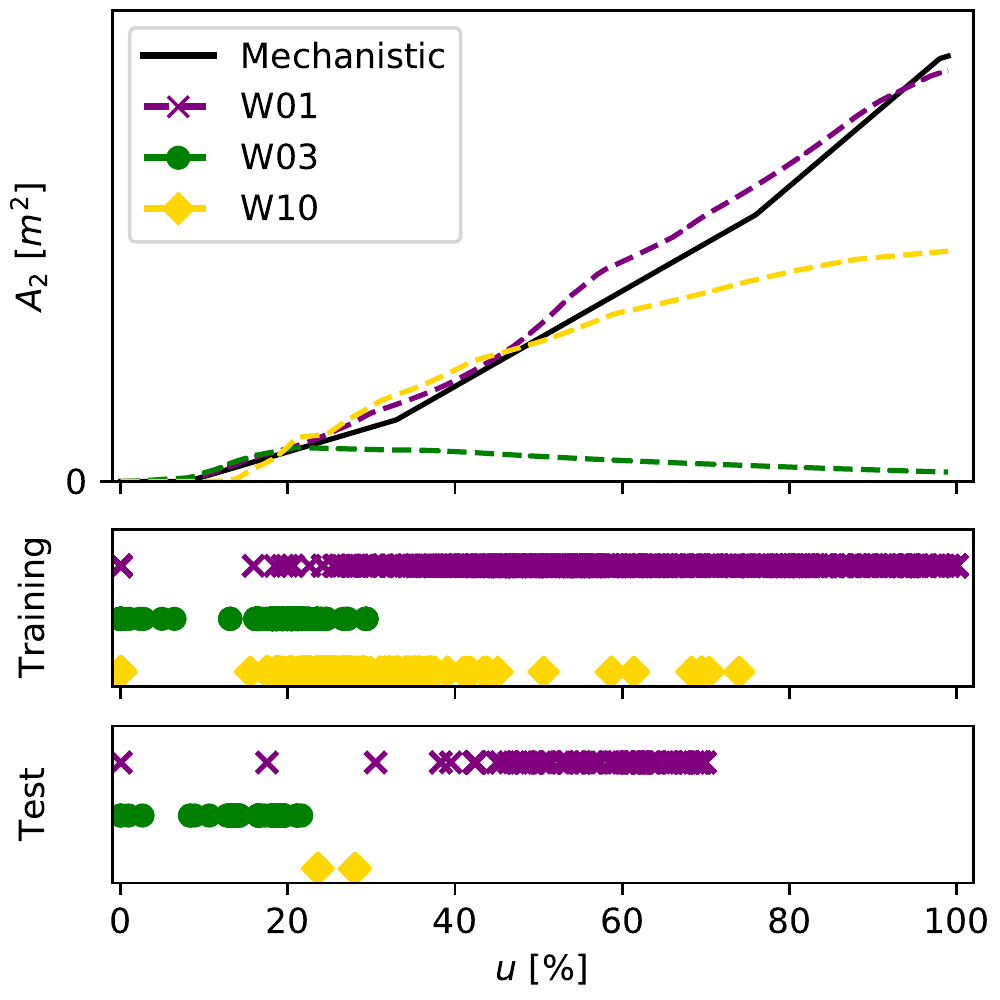}
\label{fig:eg-nn-cda2}
}
\subfloat[HM($\rho_{G,1}$)]{\includegraphics[width=0.48\textwidth]{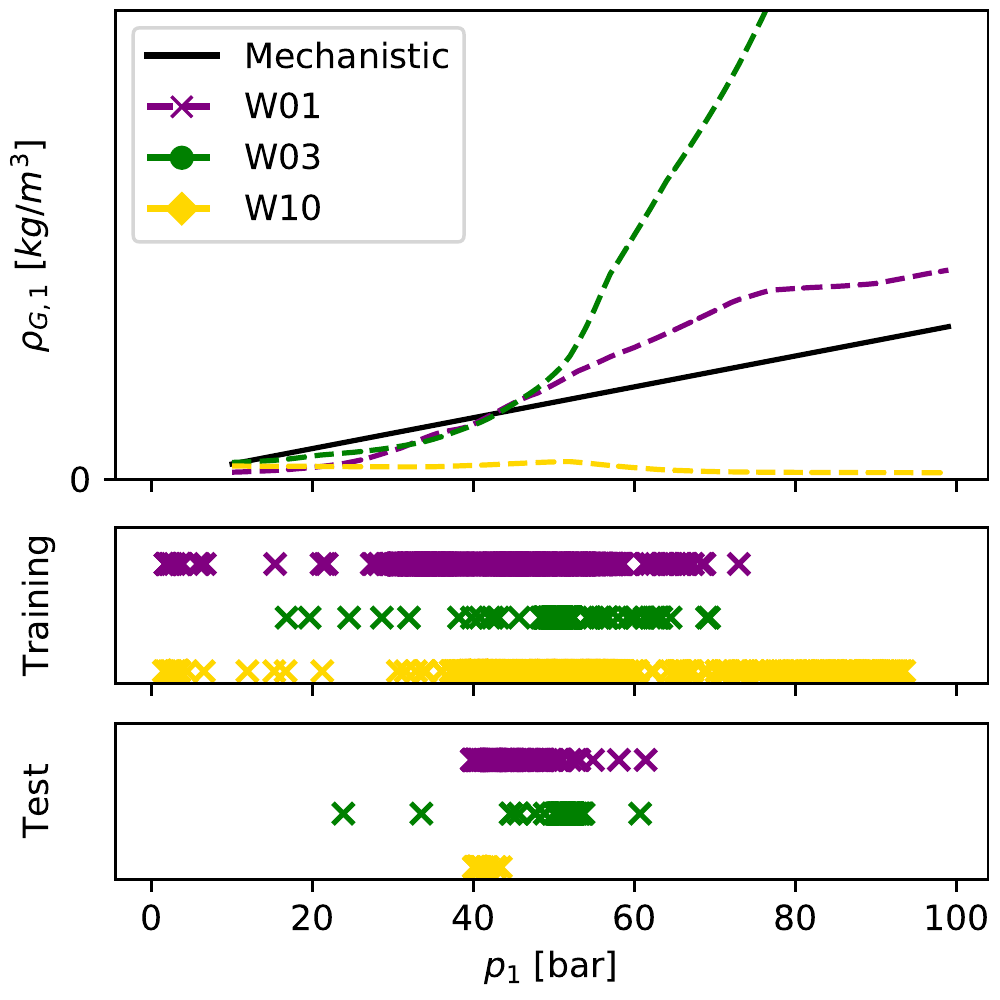}
\label{fig:eg-nn-rhoG1}
}
\caption{The learned neural network area function in (a) H($A_2$) (b)  H($\rho_{G,1}$), and for three of the wells, illustrated together with a typical mechanistic curve (black, solid). Also shown are the training and test data points for each well.}
\label{fig:eg-nn}
\end{figure}

One consideration of a model is the performance in terms of accuracy, another is the scientific consistency. Inconsistent physical behaviors may cast doubt about the trustworthiness of the models and cause the generalization abilities to be poor. First, the outputs from the neural networks in the hybrid models are investigated. Figures \ref{fig:eg-nn-cda2} and \ref{fig:eg-nn-rhoG1} shows the output from the neural network in HM($A_2$) and HM($\rho_{G,1}$), respectively, as a function of one of the inputs, for three of the wells. The results are diverse. In some of the choke models, the output of the neural network has a trend coherent with the expected physical behavior, illustrated with the mechanistic relation. This is seen for W01. However, notice that some of the other curves go to zero or explode, illustrating scientific inconsistency. This effect has also been observed for the HM($\rho_{G,2}$) and the HM($\rho$). There are two likely explanations for the nonphysical behaviors. Firstly, the behavior may be influenced by the lack of data or erroneous data. For instance, for W03, data are lacking for choke openings greater than 40\%. Secondly, due to the high capacity of neural networks, the data-driven part of a hybrid model may capture any modeling error and not just the factor or term the network was intended to represent. For instance, even though the HM($A_2$) had the best performance for W03 of all models, the area function is not in line with the expected physical behavior. This indicates that the learned neural network area function may have captured other modeling errors than just a poorly calibrated area function. 

\begin{figure}[ht]
\centering
\includegraphics[width=\textwidth]{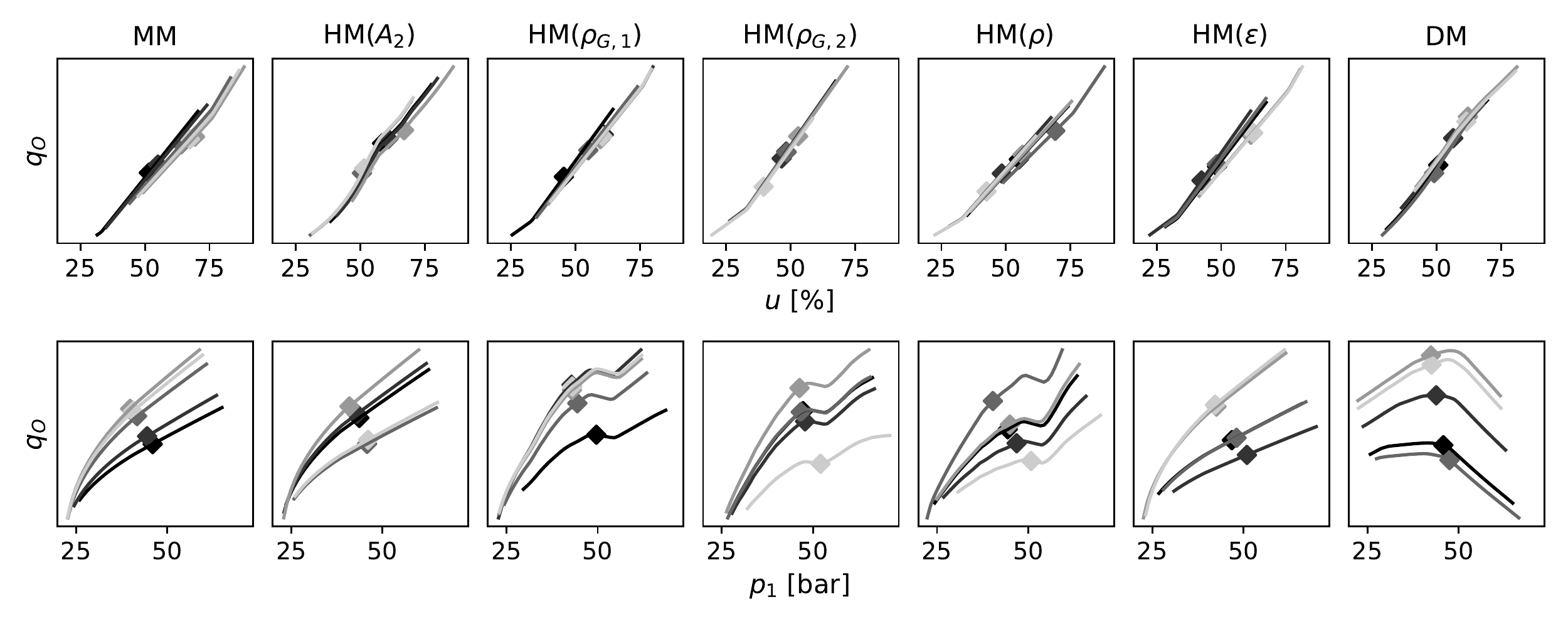}
\caption{Sensitivity analysis of the different models for W01. Five initial points are picked at random, marked with diamond, and the response of the volumetric oil flow rate when perturbing the choke opening $u$ (upper) and the upstream pressure $p_1$ (lower) is illustrated.}
\label{fig:eg-sens-W01}
\end{figure}

Additionally, a short sensitivity study is conducted to investigate the scientific consistency of the output of the seven implemented VFM models. The choke models trained on data from W01 are examined for which all models achieved a good performance, see Table \ref{tb:EG-error}. Five test points are randomly picked from the test dataset, the choke opening $u$ and the upstream pressure $p_1$ are individually perturbed and the responses in the oil volumetric flow rate $q_O$ are investigated. Under the assumption of constant process conditions and considering the production choke as an isolated unit without the influence of the rest of the production system, the oil flow rate should be expected to 1) increase with increasing choke opening, and 2) increase with increasing upstream pressure. The sensitivity study is presented in Figure \ref{fig:eg-sens-W01}. 

\begin{figure}[ht]
\centering
\includegraphics[width=0.6\textwidth]{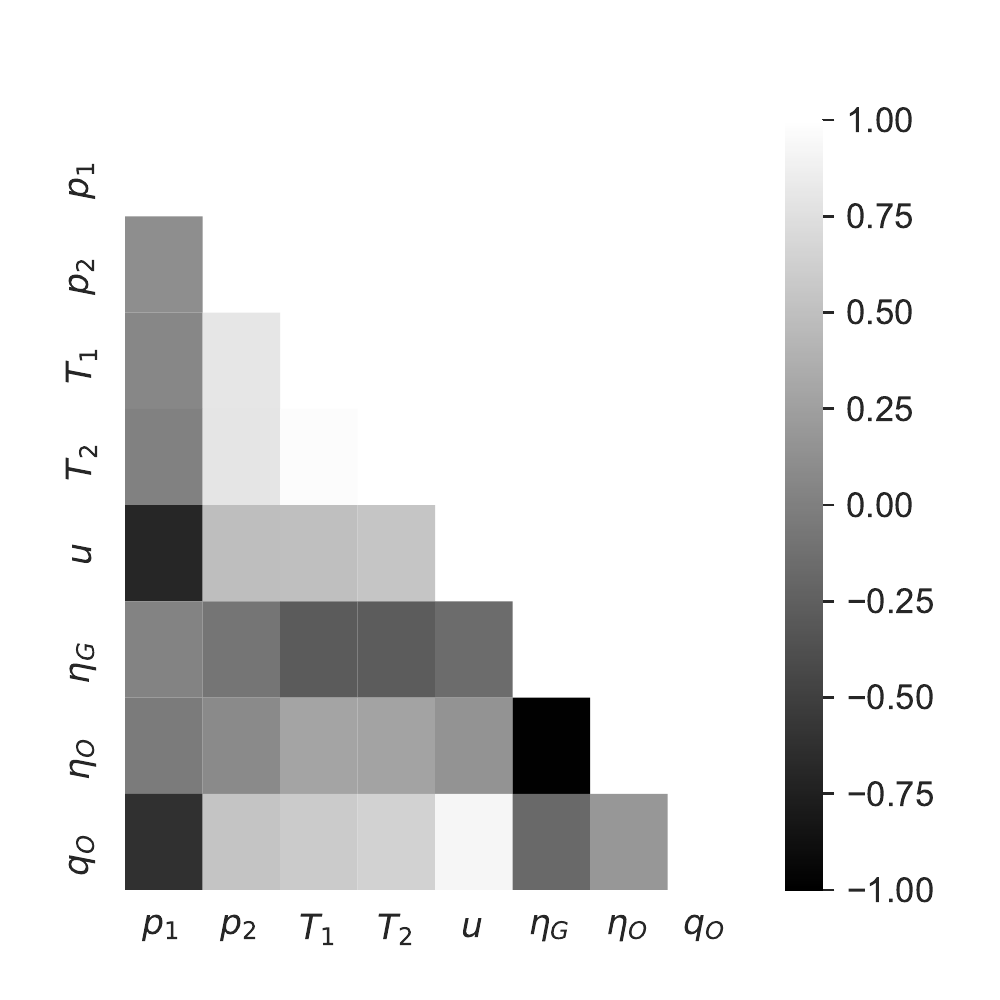}
\caption{A visualization of the correlation between the explanatory variables and the target variable measurements in the dataset corresponding to W01.}
\label{fig:correlation}
\end{figure}

Most of the models seem to mimic the expected physical behavior except for the DM, for which the oil flow rate decreases with increased pressure above a certain threshold. This effect is caused by the DM being influenced by the available data to a larger degree than the other model types, and that the available data reflects the behavior of the complete production system and not only the choke. This can be explained in more detail by looking at the correlation plot of the available measurements in the dataset corresponding to W01, see Figure \ref{fig:correlation}. Observe the negative correlation between the oil flow rate $q_O$ and the upstream pressure $p_1$. By looking at the choke as an isolated unit this correlation contradicts the expected physical behavior. On the other hand, additionally considering the wellbore, the observed correlation has a scientific explanation: increased pressure in the wellhead may result in a decreased pressure drop in the wellbore and a decreased oil flow rate. Nevertheless, if the goal of the modeling was to develop a choke model, the DM would be considered scientifically inconsistent. These results reflect upon both the positive and negative nature of models with high flexibility. They may adapt to any behavior seen in the available data, thus also erroneous data. On the other side, this sensitivity study is small and only conducted for one well. Conclusions on the scientific consistency of the general gray-box model cannot be made. Nevertheless, the results motivate the use of gray-box VFM models if scientific consistency is of importance to the end-users of the models. 

\subsection{Suggestions for improvements in future work}\label{sec:case-study-improvements}
From the results presented in Section \ref{sec:case-study} there are several aspects that can be investigated to improve upon both the prediction accuracy and the scientific consistency of hybrid models in future work. 

Firstly, only a few simplifications and assumptions are investigated as hybridization options in Section \ref{sec:models-hybridization} although numerous exist. It is likely that other hybrid model types may be better at balancing the task between learning from physics and learning from data. Further, different types of data-driven models or other mechanistic choke models may yield better performances for these wells. There is also the question raised in Section \ref{sec:case-study-predictive-performance-model-simpifications} on the suitability of different hybrid models in different cases. One approach in this direction is to utilize an advanced simulator to generate synthetic data, in which process conditions and other characteristics can be controlled. 

Secondly, Section \ref{sec:case-study-predictive-performance-data} discussed the influence of the available data on the prediction accuracy and pointed out noisy and biased measurements, together with nonstationary process conditions as influential factors. A future research path is to experiment with different estimation methods or model types that exploits knowledge regarding the uncertainty in parameters and measurements. Some examples are variational inference as estimation method, state estimation techniques such as the Kalman Filter \citep{Kalman1960}, or probabilistic models. In case of nonstationary process conditions, time dependent models may be utilized. Yet, such models greatly increase the computational complexity and may not be suitable for real-time applications. Another possibility is online learning, a learning method that may improve upon future predictive performance without adding complexity to the models.

Lastly, in Section \ref{sec:case-study-scientific-consistency}, the scientific consistency of the gray-box models were discussed and several issues raised. Several possible approaches may be investigated to improve upon the scientific consistency. Firstly, a stronger regularization of the priors obtained from the pretrained neural networks could possibly result in the network replicating the mechanistic relation to a higher degree, whilst avoiding capturing other modeling errors. Secondly, the inclusion of additional data-driven elements in a gray-box model, for instance, an error term, could enable the original data-driven element to capture the proposed physics only. Thirdly, the utilization of methods that enables learning from datasets across wells, for instance transfer learning or multitask learning, may positively change the results as more data are exploited.

\section{Concluding remarks}
This article contributes towards the development of gray-box virtual flow meters in the petroleum industry. The focus has been on white-to-gray box models where a mechanistic model is used as a baseline and data-driven elements inserted to increase model flexibility. The choke valve of 10 petroleum wells has been modeled using real production data spanning at most four years of production. 

The results are diverse with a prediction accuracy is in the range of 1.8\%-40.6\%, and no recommendations towards the suitability of different gray-box models may be drawn. The results cast light upon the nontrivial task of balancing learning from both physics and data. It is believed that the accuracy is strongly influenced by nonstationarity in the available data. Nevertheless, the results indicate that gray-box models may outperform a mechanistic and a data-driven model if an appropriate balance between the model components is identified. In particular, the gray-box modeling approach seems to increase the accuracy compared to mechanistic models and may improve the scientific consistency compared to data-driven models. 

While the gray-box modeling approaches are tested on 10 different wells, these wells, while being fairly typical offshore wells, are hardly representative for all wells. Therefore, a direct generalization of the results to other assets is difficult. Assuredly, the results could benefit from a deeper analysis of gray-box modeling on wells with significantly different characteristics. Furthermore, the research has studied the approach with VFM as application, and generalization to other application areas is inadmissible without further experimentation. On the other side, the gray-box modeling approach itself should apply to any process systems where both physical equations and process data exist. 

To this end, the results reported in this study are promising, albeit, the true potential of gray-box modeling is yet to be discovered. For example, hybrid modeling could yield great potential in the small data regime, where data-driven models are known to struggle. Several interesting research directions exist for future consideration. Among these are online learning and multi-task learning. 

\section*{Acknowledgment and funding}
This research is a part of BRU21 - NTNU Research and Innovation Program on Digital and Automation Solutions for the Oil and Gas Industry (\url{www.ntnu.edu/bru21}) and supported by Lundin Energy Norway. Lundin Energy Norway had no part in data collection and analysis, nor in writing of the report. Yet, they approved the paper before submission for publication. 

The authors would like to thank Lundin Energy Norway for the opportunity to work with and publish results related to data from the asset Edvard Grieg. We would further like to thank Solution Seeker AS for utilization of their data processing technology, saving valuable time on data curation.  

\appendix
\section{Case study - results}\label{app:eg-results}
\setcounter{figure}{1}
\begin{figure}[ht]
\centering
\subfloat[W01]{
\includegraphics[width=0.48\textwidth]{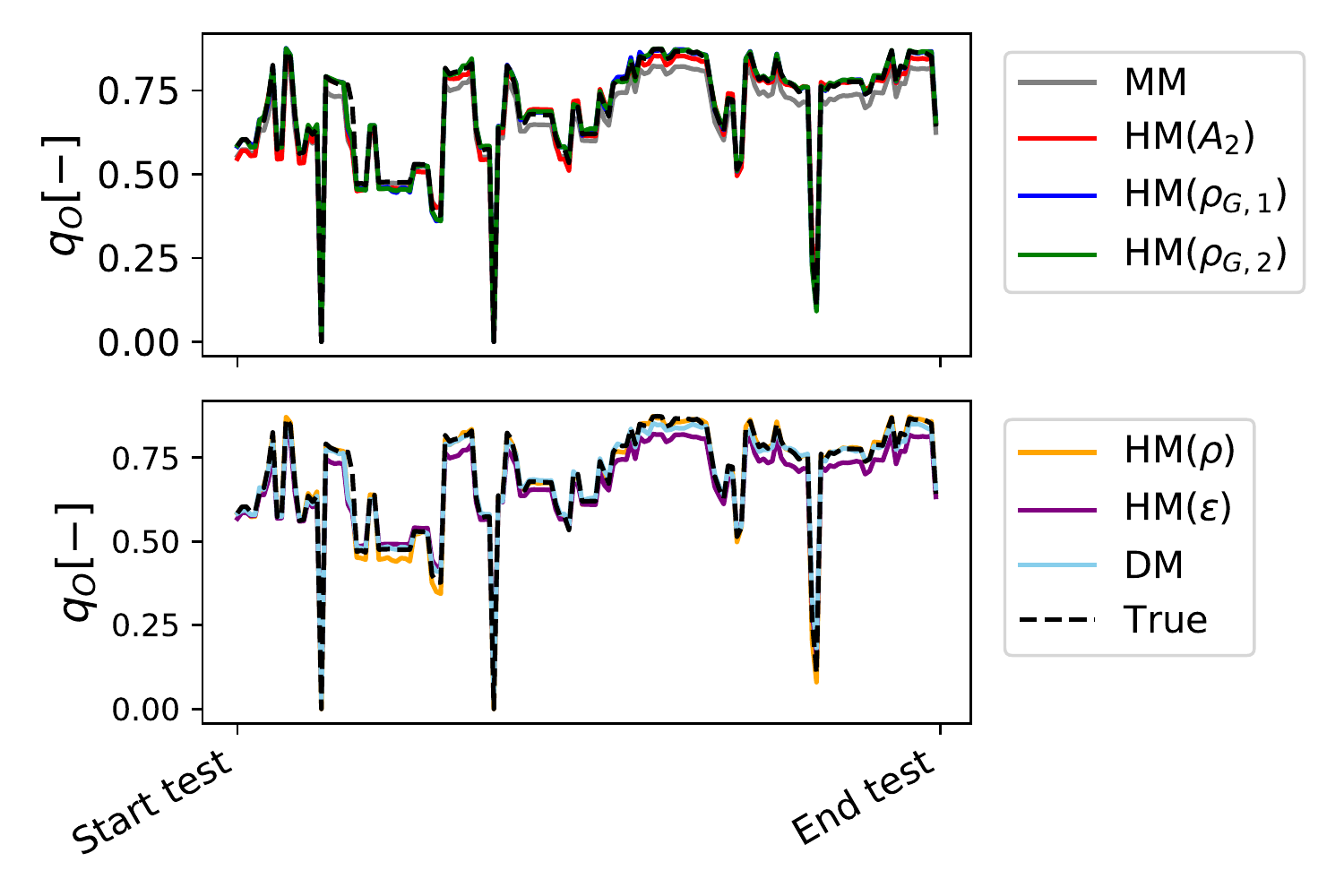}
\label{fig:flow-w1}
}
\hfill
\subfloat[W02]{
\includegraphics[width=0.48\textwidth]{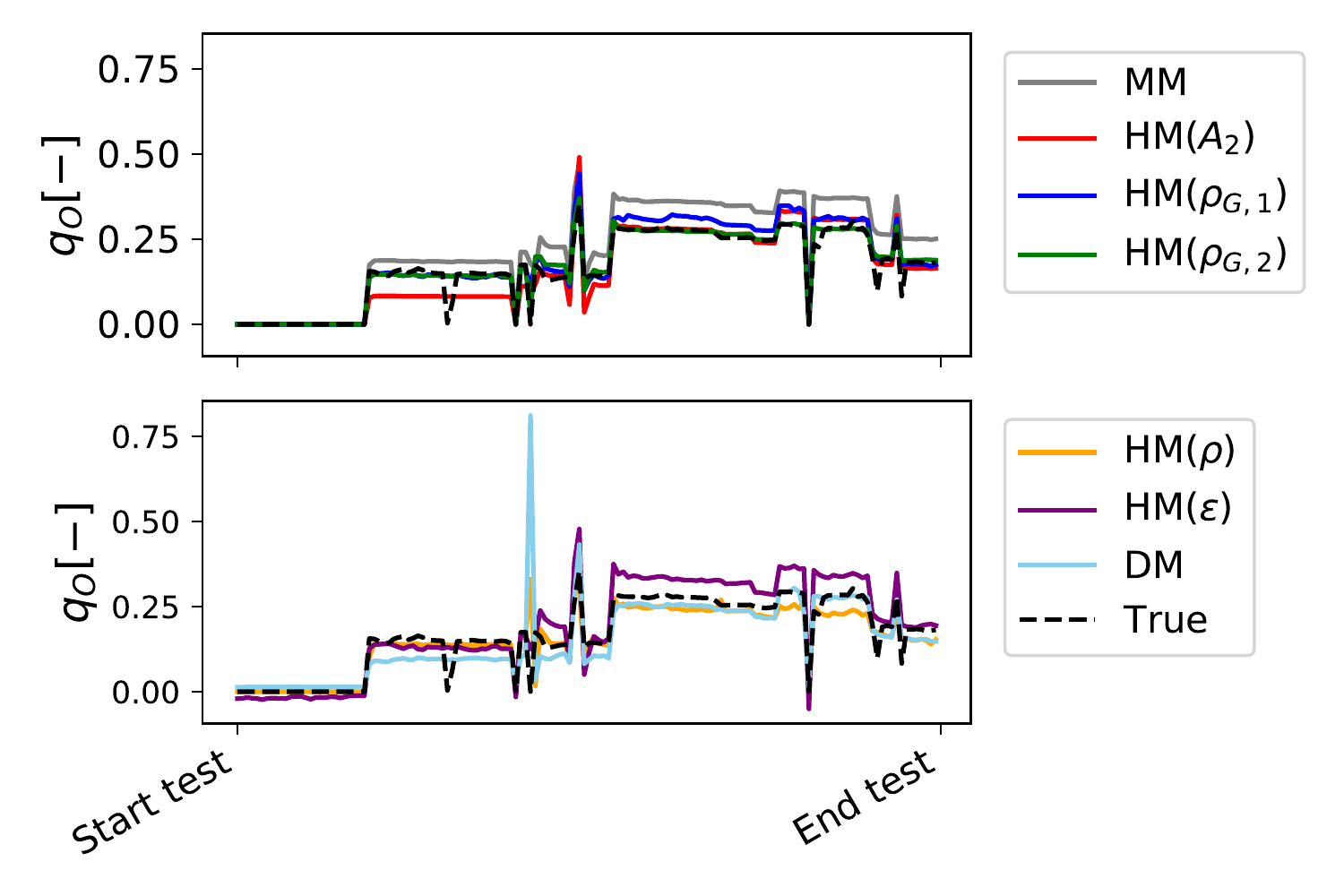}
\label{fig:flow-w2}
}
\hfill
\subfloat[W03]{
\includegraphics[width=0.48\textwidth]{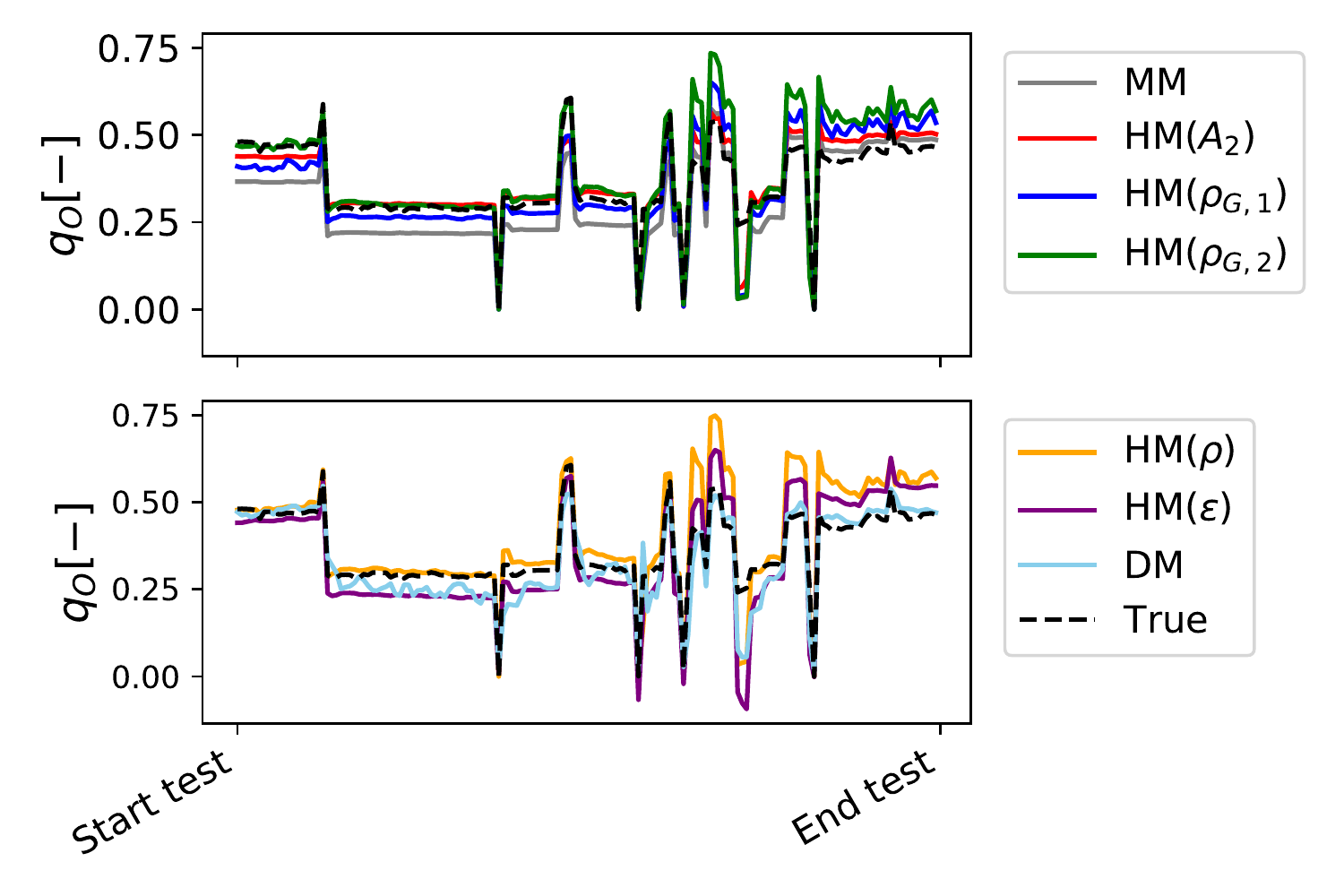}
\label{fig:flow-w3}
}
\hfill
\subfloat[W04]{
\includegraphics[width=0.48\textwidth]{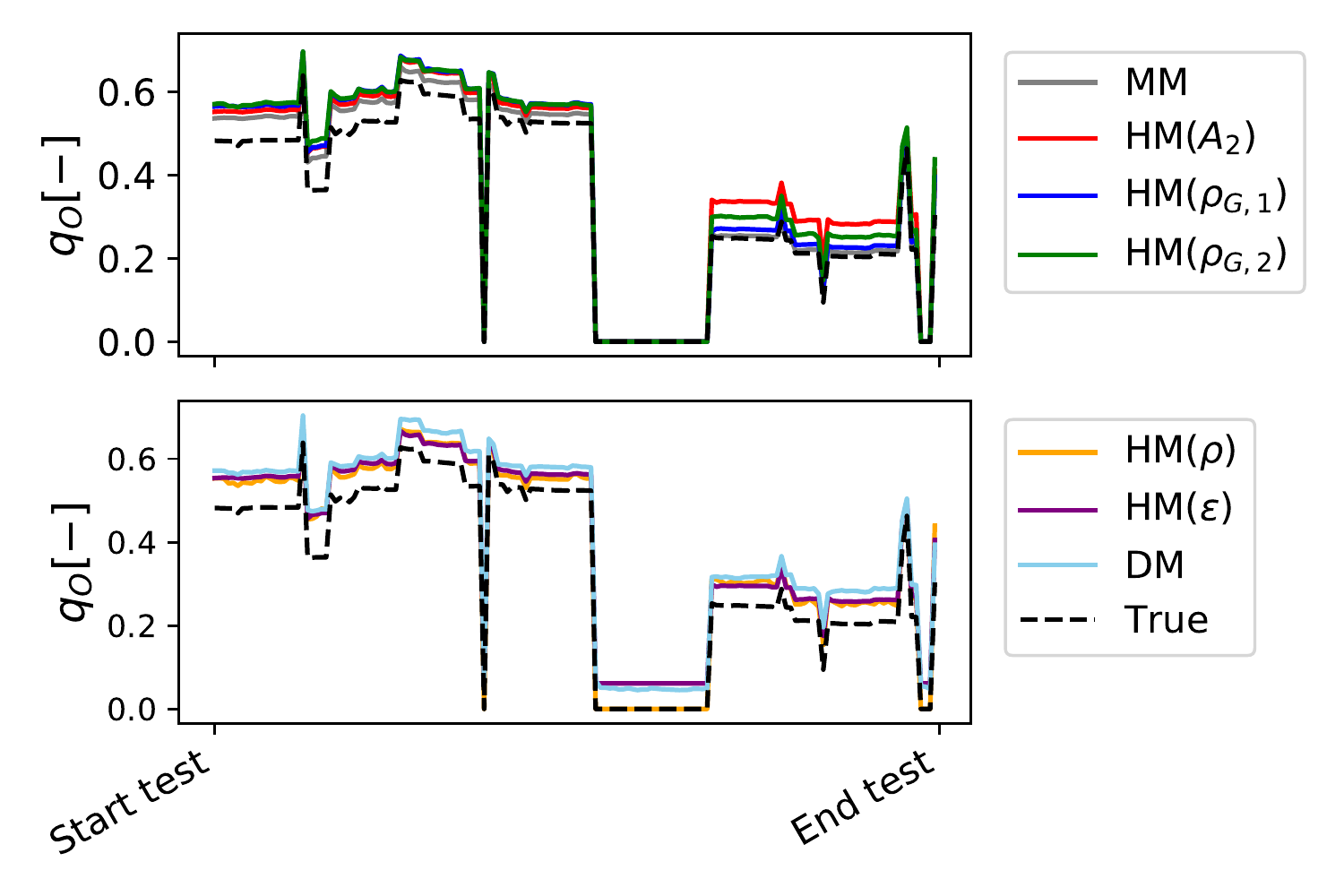}
\label{fig:flow-w4}
}
\hfill
\subfloat[W05]{
\includegraphics[width=0.48\textwidth]{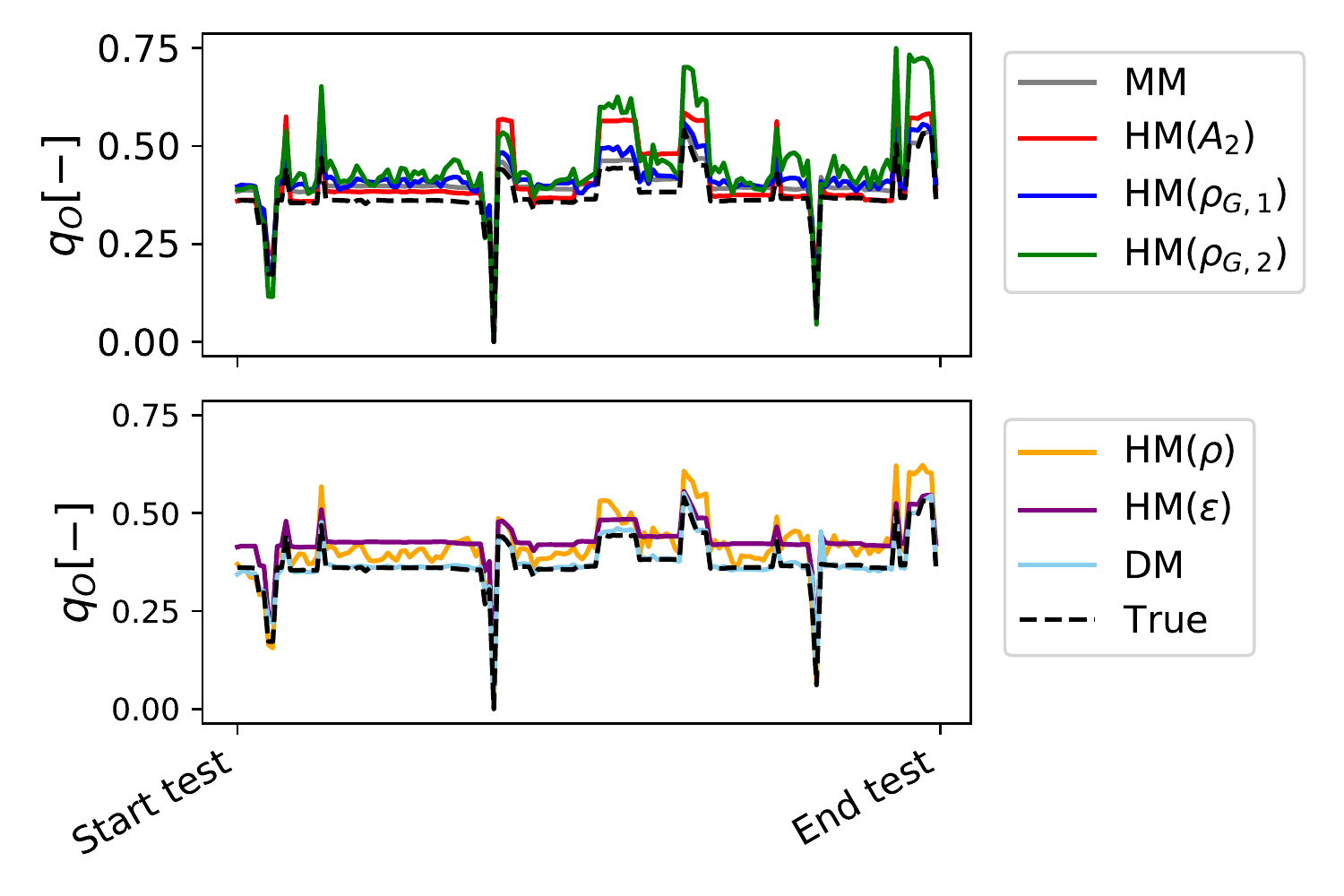}
\label{fig:flow-w5}
}
\hfill
\subfloat[W06]{
\includegraphics[width=0.48\textwidth]{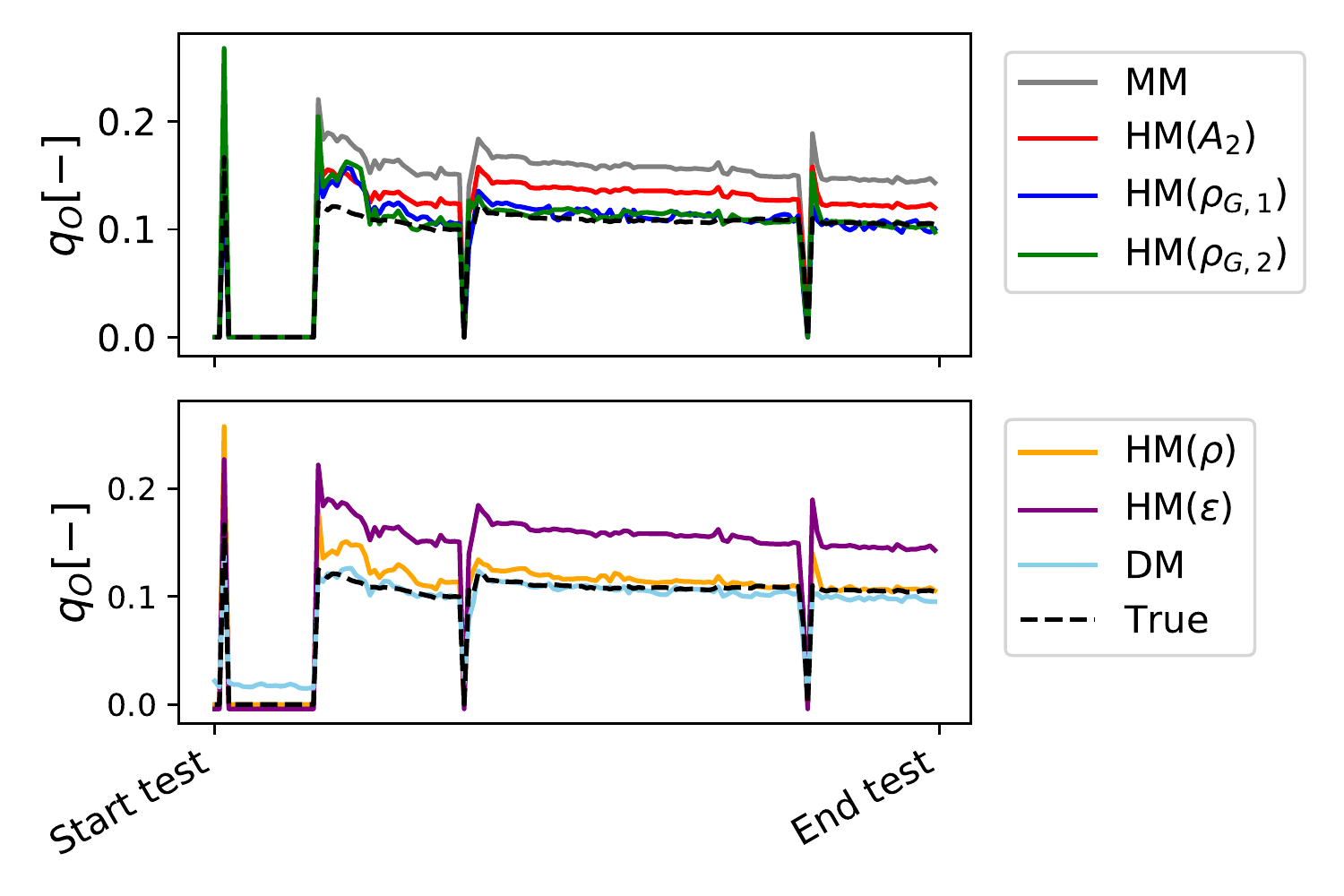}
\label{fig:flow-w6}
}
\end{figure}

\begin{figure}
\ContinuedFloat
\subfloat[W07]{
\includegraphics[width=0.48\textwidth]{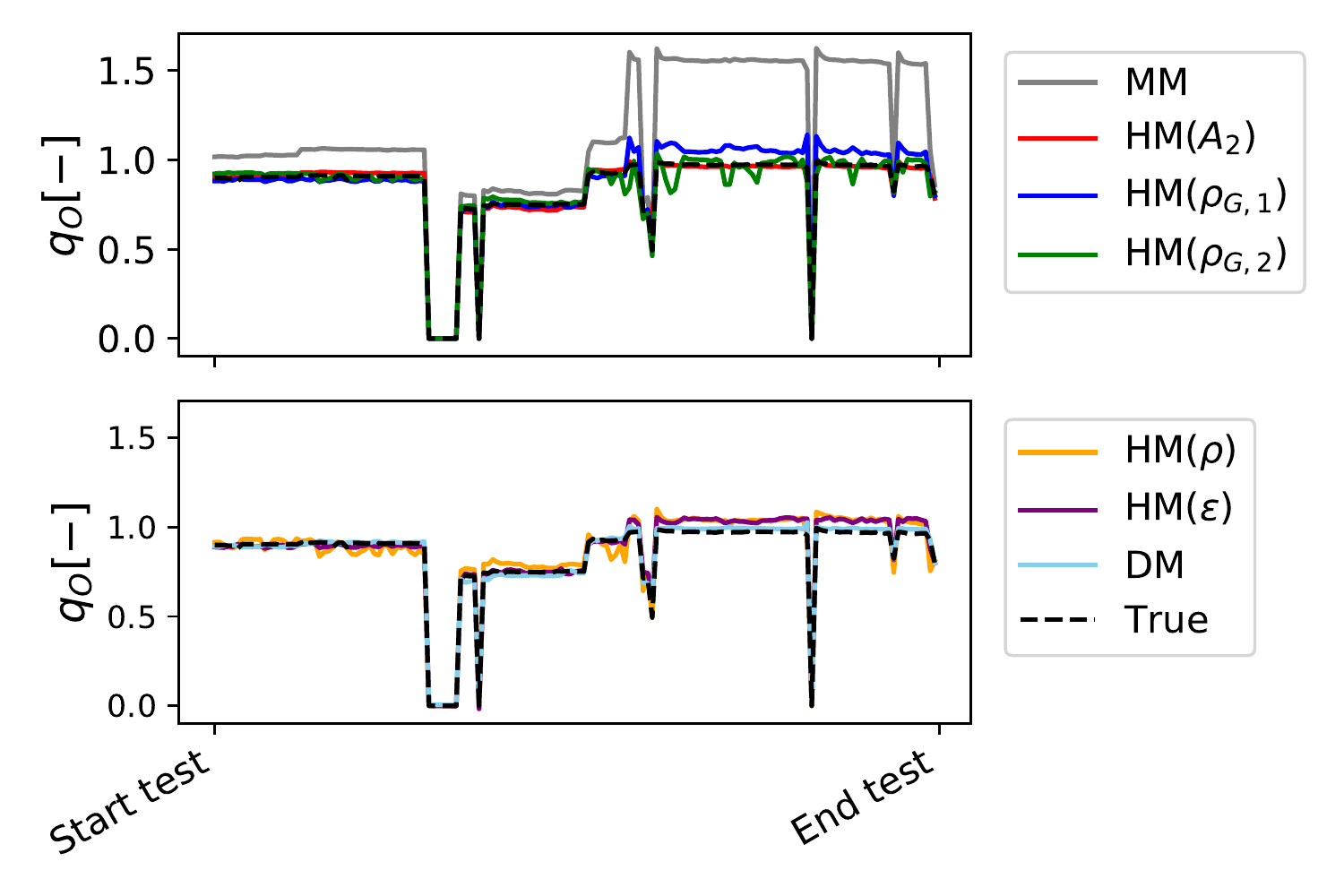}
\label{fig:flow-w7}
}
\hfill
\subfloat[W08]{
\includegraphics[width=0.48\textwidth]{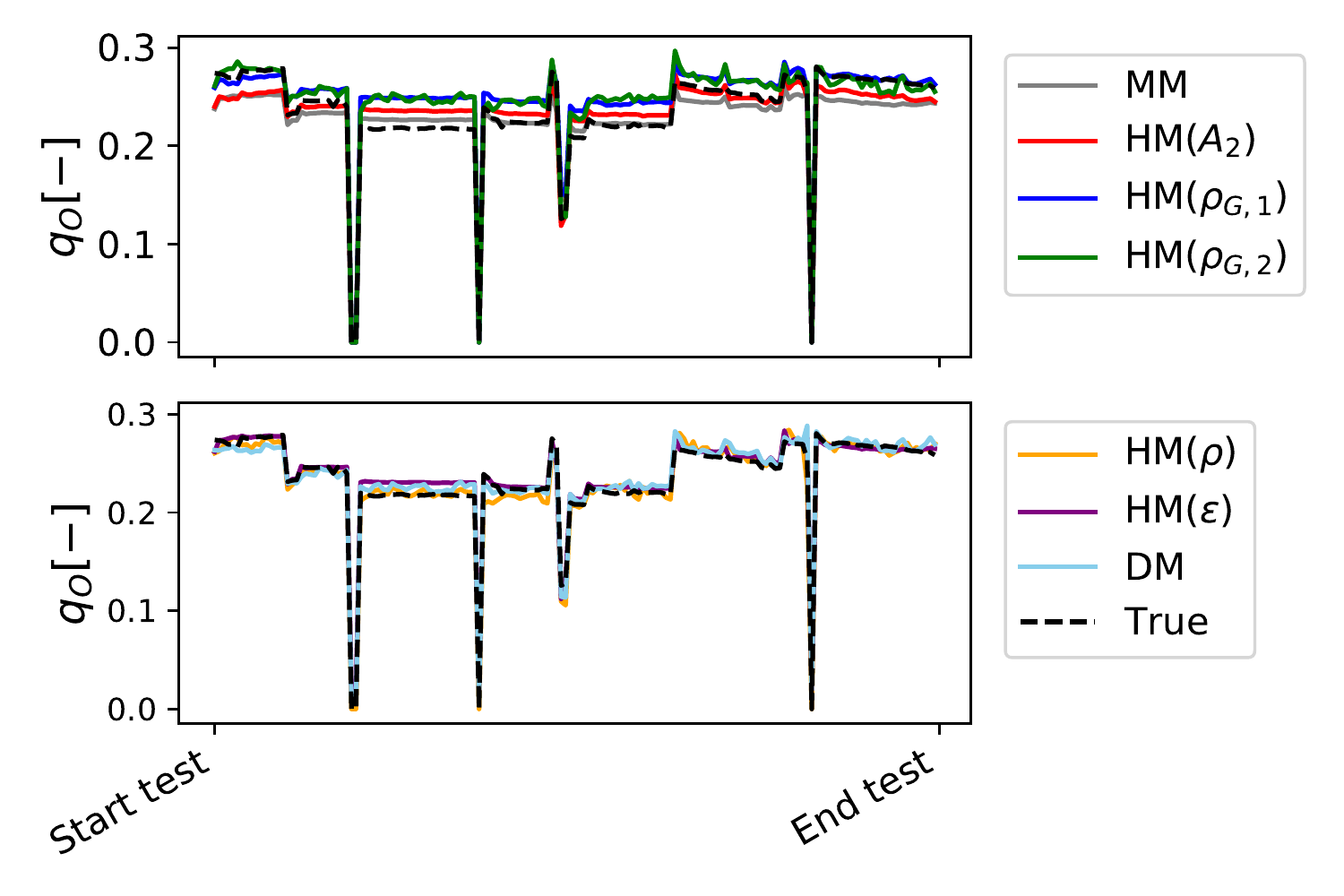}
\label{fig:flow-w8}
}
\hfill
\subfloat[W09]{
\includegraphics[width=0.48\textwidth]{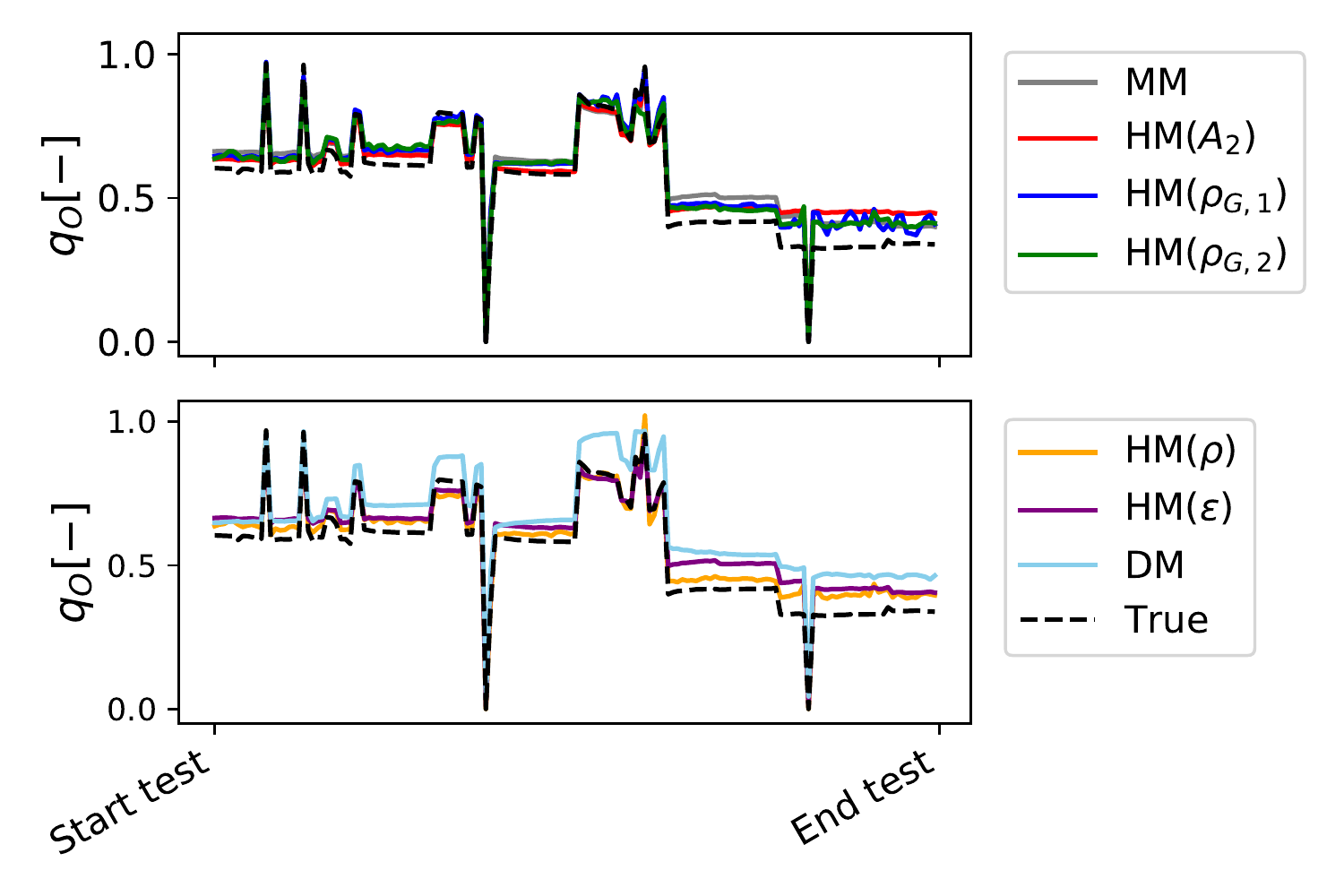}
\label{fig:flow-w9}
}
\hfill
\subfloat[W10]{
\includegraphics[width=0.48\textwidth]{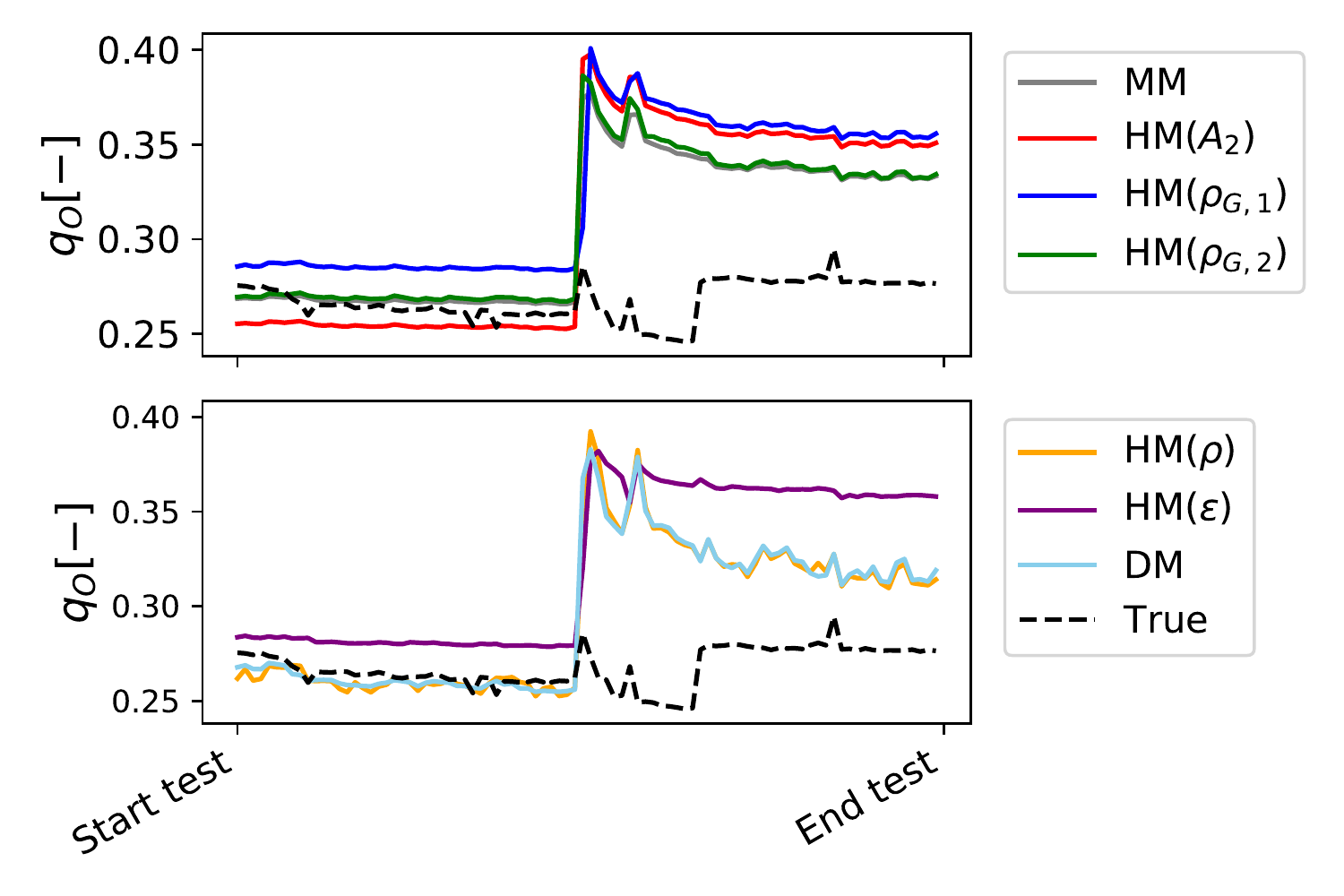}
\label{fig:flow-w10}
}
\caption{Illustration of the (downscaled) volumetric oil flow rate for each of the well and all models. Shown in dotted black are the measured flow rate from the multiphase flow meter. Notice, for some of the wells all models have adequate prediction accuracy, whilst for other wells, some model predictions are unsatisfactory.}
\label{fig:app-eg-flow}
\end{figure}

\begin{figure}[h!]
\centering
\subfloat[MM]{
\includegraphics[width=0.47\textwidth]{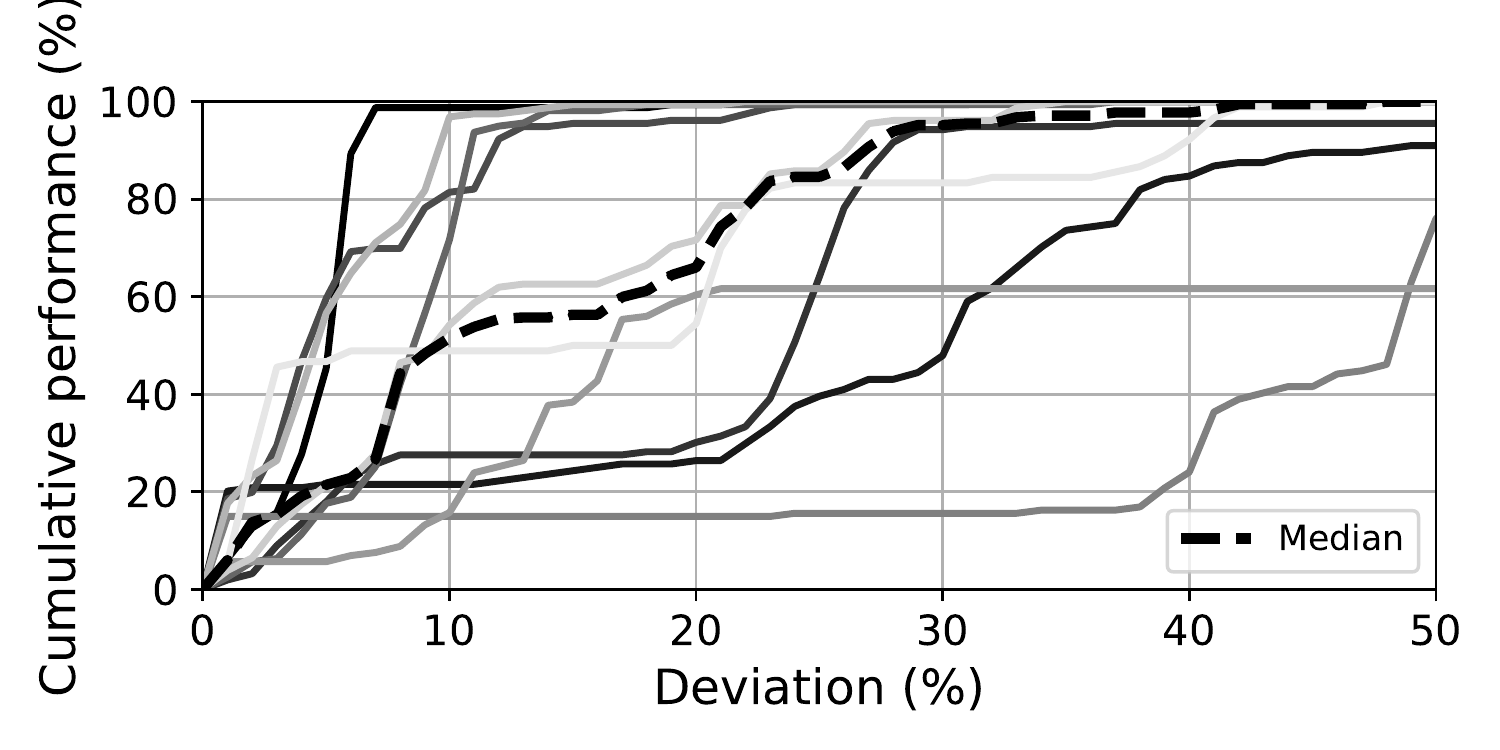}
\label{fig:cum-perf-mm}
}
\hfill
\subfloat[HM($A_2$)]{
\includegraphics[width=0.47\textwidth]{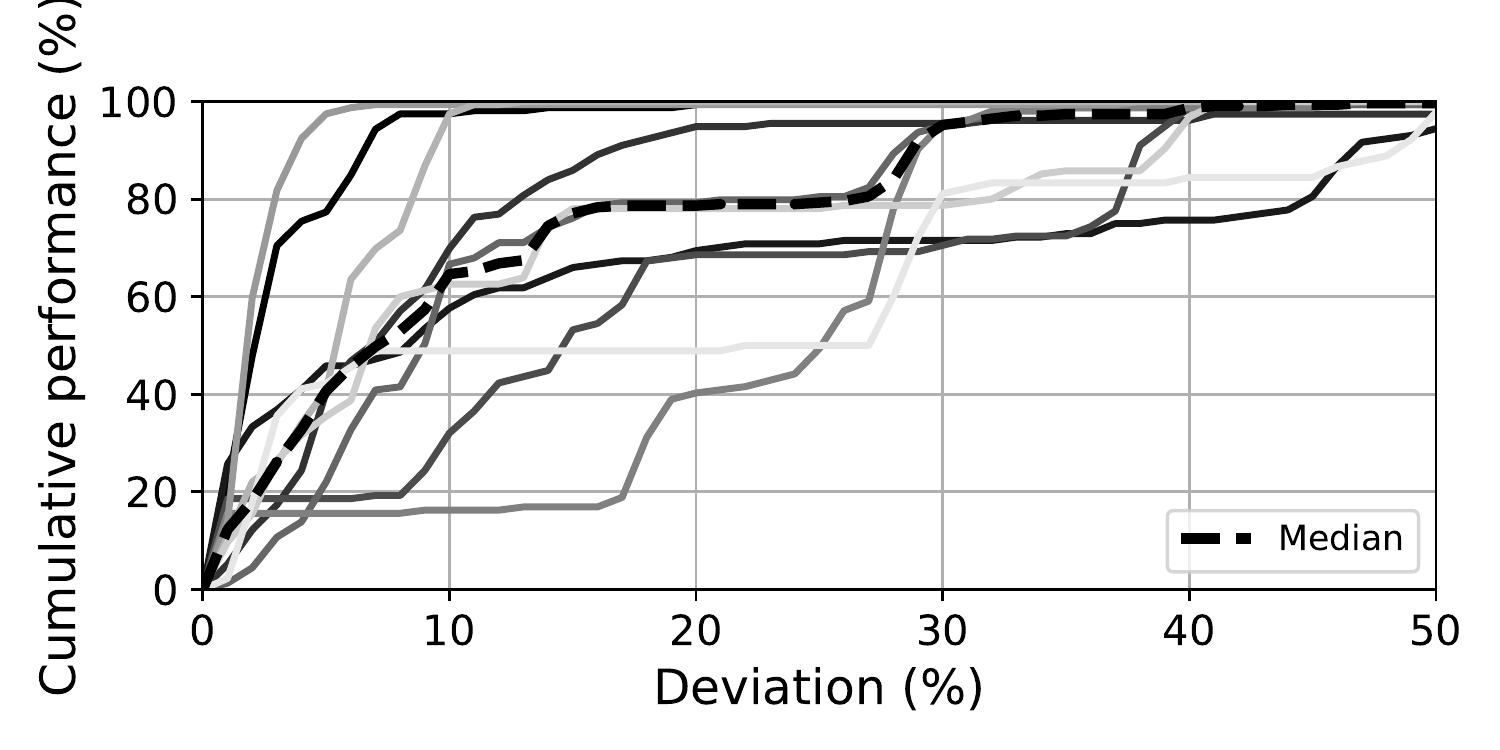}
\label{fig:cum-perf-cda2}
}
\hfill
\subfloat[HM($\rho_{G,1}$)]{
\includegraphics[width=0.47\textwidth]{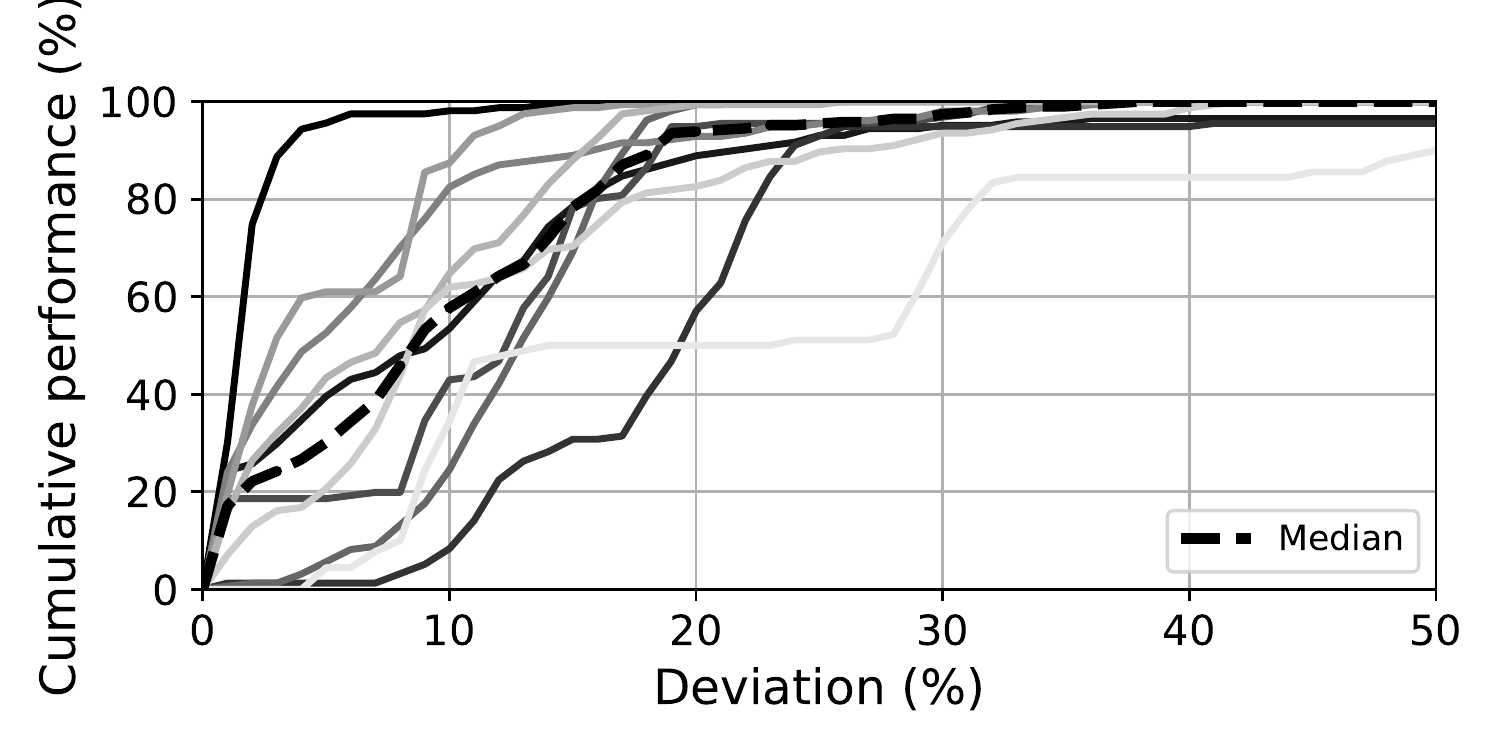}
\label{fig:cum-perf-rho-g1}
}
\hfill
\subfloat[HM($\rho_{G,2}$)]{
\includegraphics[width=0.47\textwidth]{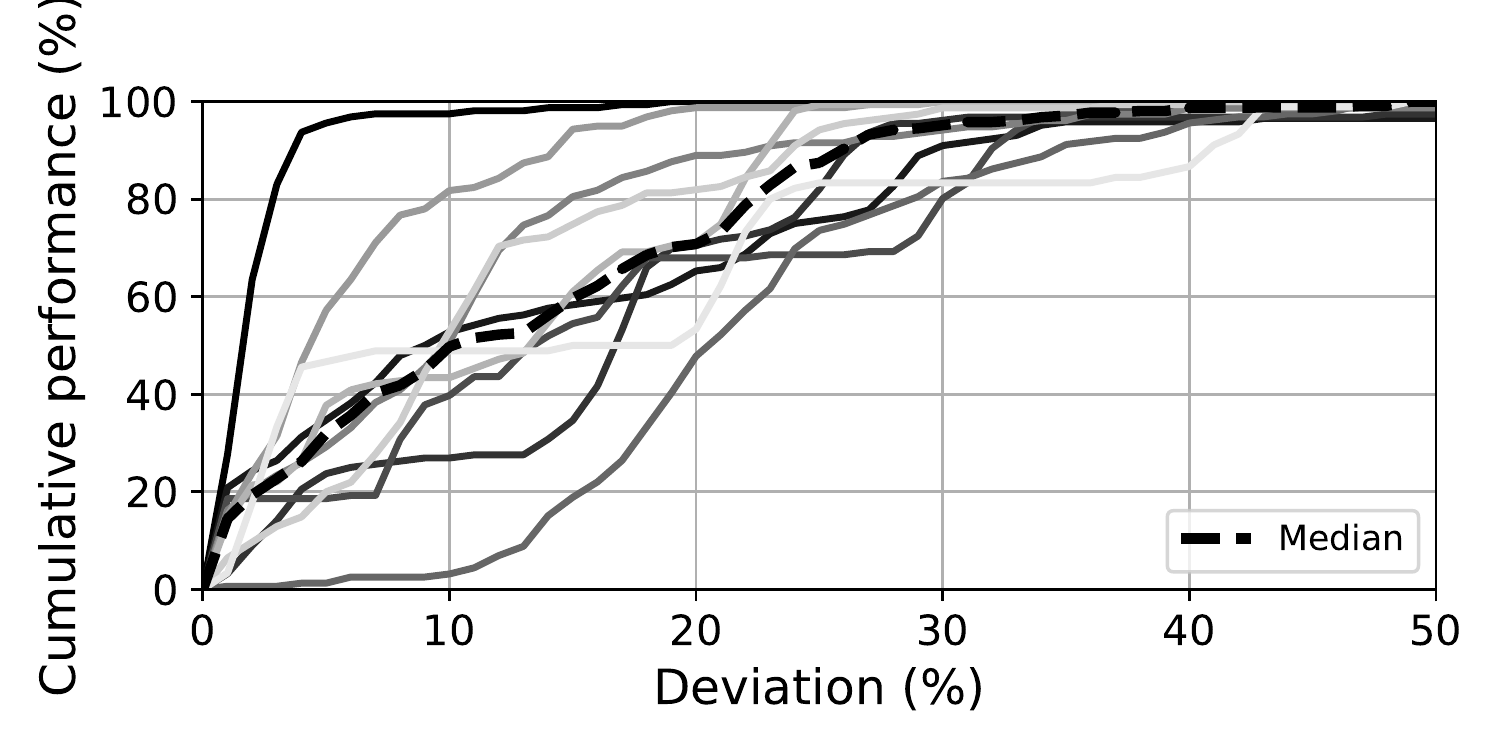}
\label{fig:cum-perf-rho-g2}
}
\hfill
\subfloat[HM($\rho$)]{
\includegraphics[width=0.47\textwidth]{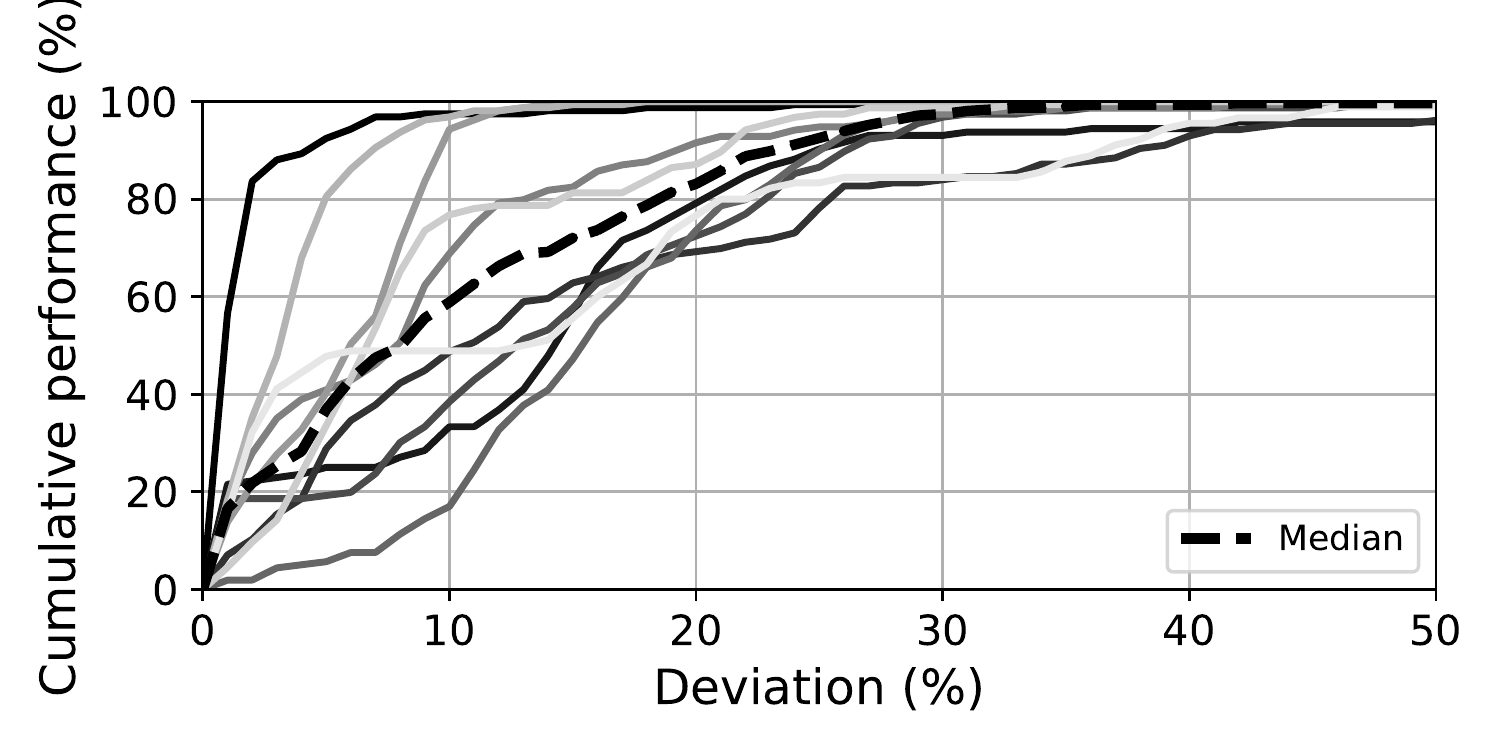}
\label{fig:cum-perf-rho}
}
\hfill
\subfloat[HM($\varepsilon$)]{
\includegraphics[width=0.47\textwidth]{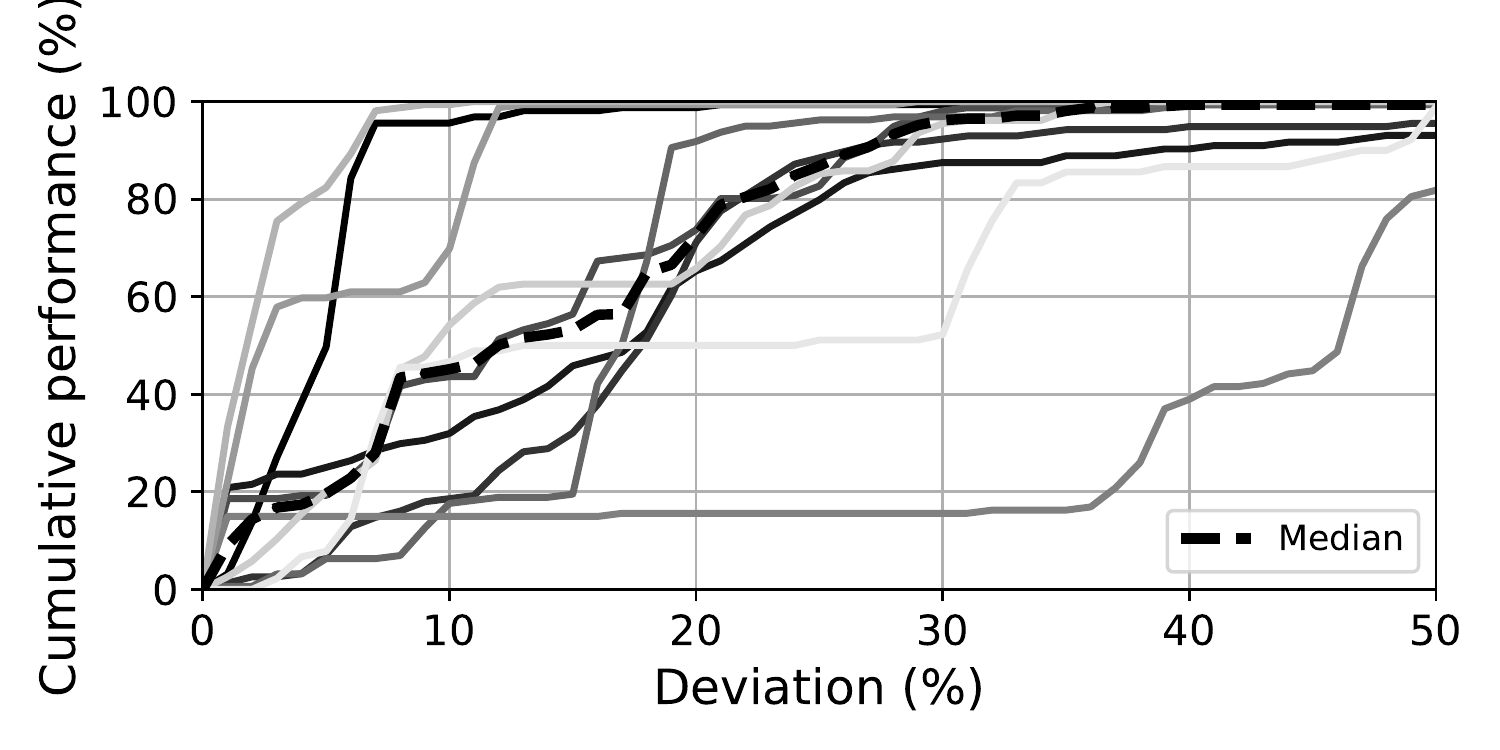}
\label{fig:cum-perf-corr}
}
\hfill
\subfloat[DM]{
\includegraphics[width=0.47\textwidth]{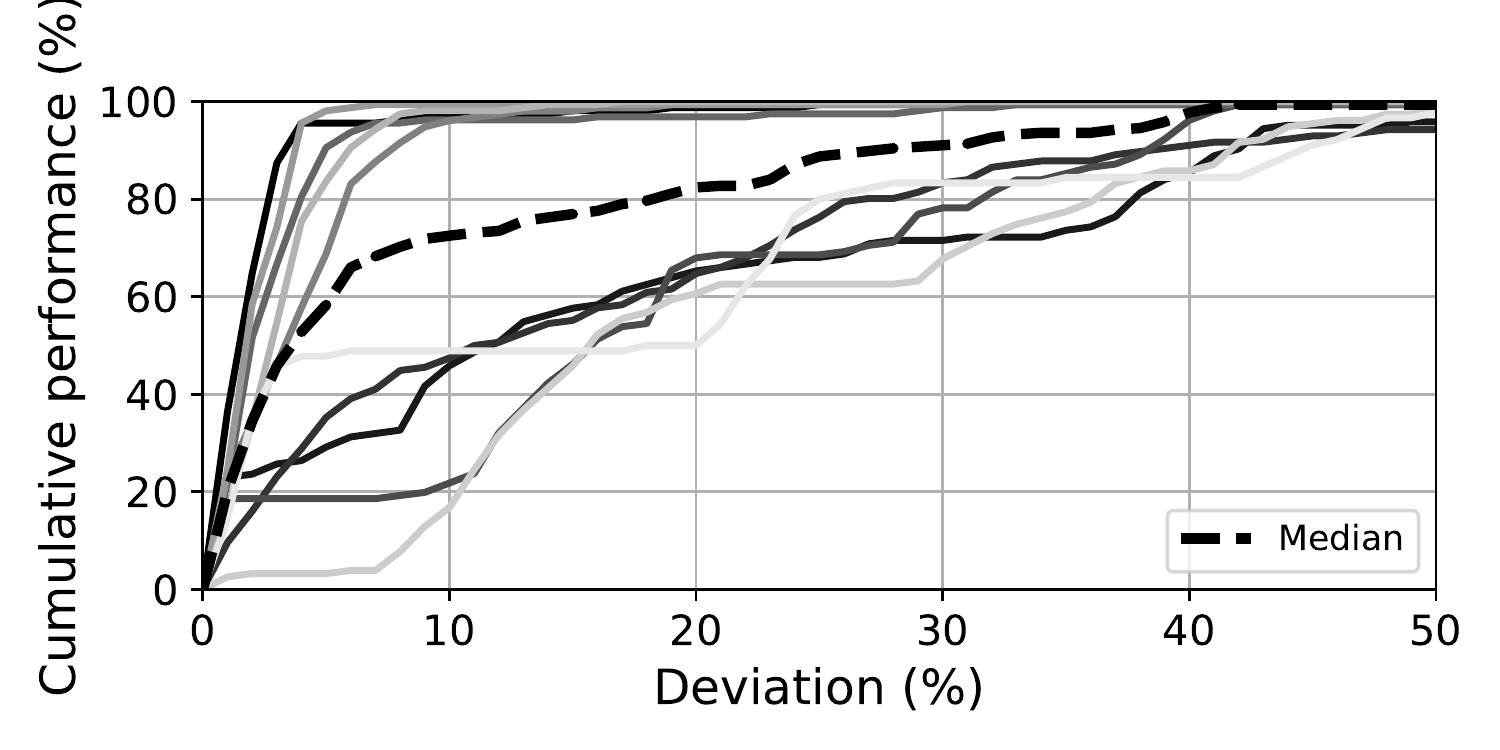}
\label{fig:cum-perf-dm}
}
\caption{Cumulative performance of choke models grouped on the model types. The black dotted line shows the median performance across wells.}
\label{fig:eg-cum-perf-all-wells}
\end{figure}

\clearpage 
\bibliography{ms}

\end{document}